\documentclass[pdflatex,sn-mathphys-num]{sn-jnl}%

\makeatletter
\@twosidefalse
\@mparswitchfalse
\makeatother
\usepackage{graphicx}%
\usepackage{multirow}%
\usepackage{amsmath,amssymb,amsfonts}%
\usepackage{amsthm}%
\usepackage{mathrsfs}%
\usepackage[title]{appendix}%
\usepackage{xcolor}%
\usepackage{textcomp}%
\usepackage{manyfoot}%
\usepackage{booktabs}%
\usepackage{algorithm}%
\usepackage{algorithmicx}%
\usepackage{algpseudocode}%
\usepackage{listings}%
\usepackage{bibunits}
\usepackage{array}
\usepackage{url}
\usepackage{placeins}

\theoremstyle{thmstyleone}%

\theoremstyle{thmstyletwo}%

\theoremstyle{thmstylethree}%

\begin{document}

\title[Article Title]{NeuralCrop: Combining physics and machine learning for improved crop modelling} %

\author*[1,2]{\fnm{Yunan} \sur{Lin}}\email{yunan.lin@tum.de}

\author[1,2]{\fnm{Sebastian} \sur{Bathiany}}

\author[1,2]{\fnm{Maha} \sur{Badri}}

\author[2,1]{\fnm{Maximilian} \sur{Gelbrecht}}

\author[1,2]{\fnm{Philipp} \sur{Hess}}

\author[2,1]{\fnm{Brian} \sur{Groenke}}

\author[2]{\fnm{Jens} \sur{Heinke}}

\author[2]{\fnm{Christoph} \sur{Müller}}

\author*[1,2,3]{\fnm{Niklas} \sur{Boers}}\email{n.boers@tum.de}

\affil[1]{\orgname{Munich Climate Center and Earth System Modelling Group, Department of Aerospace and Geodesy, School of Engineering and Design, Technical University of Munich}, \city{Munich}, \country{Germany}}

\affil[2]{\orgname{Potsdam Institute for Climate Impact Research}, \city{Potsdam}, \country{Germany}}

\affil[3]{\orgname{University of Exeter}, \city{Exeter}, \country{UK}}

\abstract{

Global gridded crop models (GGCMs) are crucial for projecting climate impacts on agricultural productivity and assessing associated risks to food security. However, state-of-the-art GGCMs are subject to substantial uncertainties arising from reduced representations of complex processes. Projected increases in the frequency and severity of climate extremes require more reliable agricultural yield simulations. Recent machine learning approaches trained on observational data provide alternatives to GGCMs, but remain under-constrained due to sparse observations and show poor out-of-distribution generalization. Here we introduce NeuralCrop, a differentiable hybrid GGCM that combines the strengths of an advanced process-based GGCM (LPJmL) with data-driven neural network components. NeuralCrop is first trained to reproduce its underlying fully process-based version, before it is fine-tuned on observational flux data of physiological variables. Evaluation against held-out flux-tower observations from European wheat and US maize sites shows 32.6--45.2\% reductions in mean normalized root-mean-square error for daily carbon and water fluxes. Although fine-tuned only on physiological variables, NeuralCrop improves temporal correlations with independent yield records across 71.4\% of the cultivated area for European wheat and 84.0\% for the US Corn Belt during 1981--2019. It performs highly competitively compared to GGCMI Phase~3 models. NeuralCrop reproduces yield anomalies more accurately across a range of soil-moisture conditions and especially for extreme drought conditions. For large-scale, long-term simulations, NeuralCrop runs nearly two orders of magnitude faster than the CPU-based LPJmL baseline model, mainly thanks to GPU acceleration. Our results highlight end-to-end hybrid crop modelling as a promising avenue for improving crop simulations and informing food security assessments under climate change.

}

\maketitle

\section*{Introduction}

Global food systems face intensifying threats from climate change, particularly extreme weather events \cite{yang2024climate, rezaei2023climate, hasegawa2021extreme}. Accurately assessing crop responses to a changing climate is crucial for developing adaptation strategies and sustaining future food security~\cite{hultgren2025impacts, lesk2016influence}. Global gridded crop models (GGCMs) are essential tools for projecting climate impacts on crop yields \cite{jagermeyr2021climate, adrian2022climate}, and are extensively used in downstream analyses of food security~\cite{Beier2025PlanetaryBoundaries, luchtenbelt2024quantifying}, sustainability \cite{kahiluoto2024redistribution}, and economic impacts \cite{orlov2024human}. Despite decades of development, state-of-the-art GGCMs remain subject to substantial uncertainties arising from imperfect process representations and model parameterizations~\cite{wang2024pathways, xiong2020different, asseng2013uncertainty, Muller2024Substanti, iizumi2020global}. Given the projected increases in frequency and intensity of extreme events in a warming climate~\cite{lee2023climate}, these uncertainties present major challenges for reliable yield projections and food security assessments~\cite{koehler2020uncertainties}.

Machine learning (ML) approaches have emerged in agricultural ecosystem modelling \cite{sweet2025transdisciplinary, liu2024knowledge, van2020crop}. These models can learn complex, nonlinear interactions from observational data (e.g., remote sensing, field measurements, and yield statistics) that are difficult to represent explicitly in process-based GGCMs, offering an alternative approach to crop yield projection. ML approaches have been successfully used to emulate GGCMs for sensitivity and uncertainty analyses \cite{liu2023statistical}, and to downscale GGCM outputs to finer spatial and temporal resolutions \cite{folberth2019spatio}. Once trained, ML approaches typically require considerably lower computational cost for predictions than traditional GGCMs.

However, ML approaches have noteworthy limitations compared with GGCMs. Without explicit representations of biophysical mechanisms, purely data-driven models primarily learn statistical relationships, which can limit interpretability and extrapolation to unseen future climates \cite{paudel2023interpretability, sweet2025transdisciplinary}. ML models trained on historical climate--yield relationships show markedly poorer performance in reproducing yield anomalies under extreme climatic conditions~\cite{beillouin2020impact}. Ensuring generalization and physical consistency thus remains a primary barrier to their application in crop modelling. These challenges are amplified by the limited availability of high-quality observational data~\cite{tseng2021learning}.
 
In light of the complementary benefits and limitations of process-based and ML-based models, hybrid approaches have been proposed as a promising direction for Earth system model development \cite{reichstein2019deep, irrgang2021towards}. By replacing or augmenting uncertain process representations with data-driven components~\cite{gelbrecht2023differentiable, shen2023differentiable}, hybrid models have shown promise across diverse Earth system applications~\cite{liu2025elm2, kochkov2024neural, zhang2023implementation}. Previous studies of ML in crop modelling include model emulation and the integration of independently trained components~\cite{von2024knowledge,droutsas2022integration,maestrini2022mixing,shahhosseini2021coupling}. However, optimizing selected trainable components jointly within a GGCMs simulation has not been achieved so far, because it would require automatic differentiation through the simulated model trajectory \cite{gelbrecht2023differentiable}. Established GGCMs are not differentiable and are thus, in particular, not designed to propagate gradients through complete simulations, making end-to-end optimization of embedded ML components difficult. Recent advances in differentiable programming make it increasingly practical to integrate and optimize trainable components within process-based simulations \cite{moses2026dj4earth}.

Here we introduce NeuralCrop, an end-to-end differentiable hybrid GGCM built on process representations of the GGCM LPJmL \cite{2018LPJmL}, and augmented with trainable ML components. By implementing the model in a differentiable form, NeuralCrop enables end-to-end online training of these components through the evolving model dynamics. Due to sparse agricultural observations, NeuralCrop is first pre-trained to reproduce simulations generated by its fully process-based version (which is essentially a Julia implementation of LPJmL) and then fine-tuned using site-level eddy-covariance observations. Across held-out site seasons, NeuralCrop improves simulations of daily carbon and water fluxes for wheat and maize. Although fine-tuned only on site-level fluxes, it also better captures interannual yield variability in the cultivated areas of European wheat and maize in the US Corn Belt, and reproduces yield anomalies more accurately across soil-moisture conditions and especially during drought conditions. These results show that fine-tuning with site-level flux observations can improve agreement with independent regional yield records. 

\section*{NeuralCrop}

\begin{figure}[!t]
\centering
\makebox[\textwidth][c]{%
\includegraphics[width=1.15\textwidth]{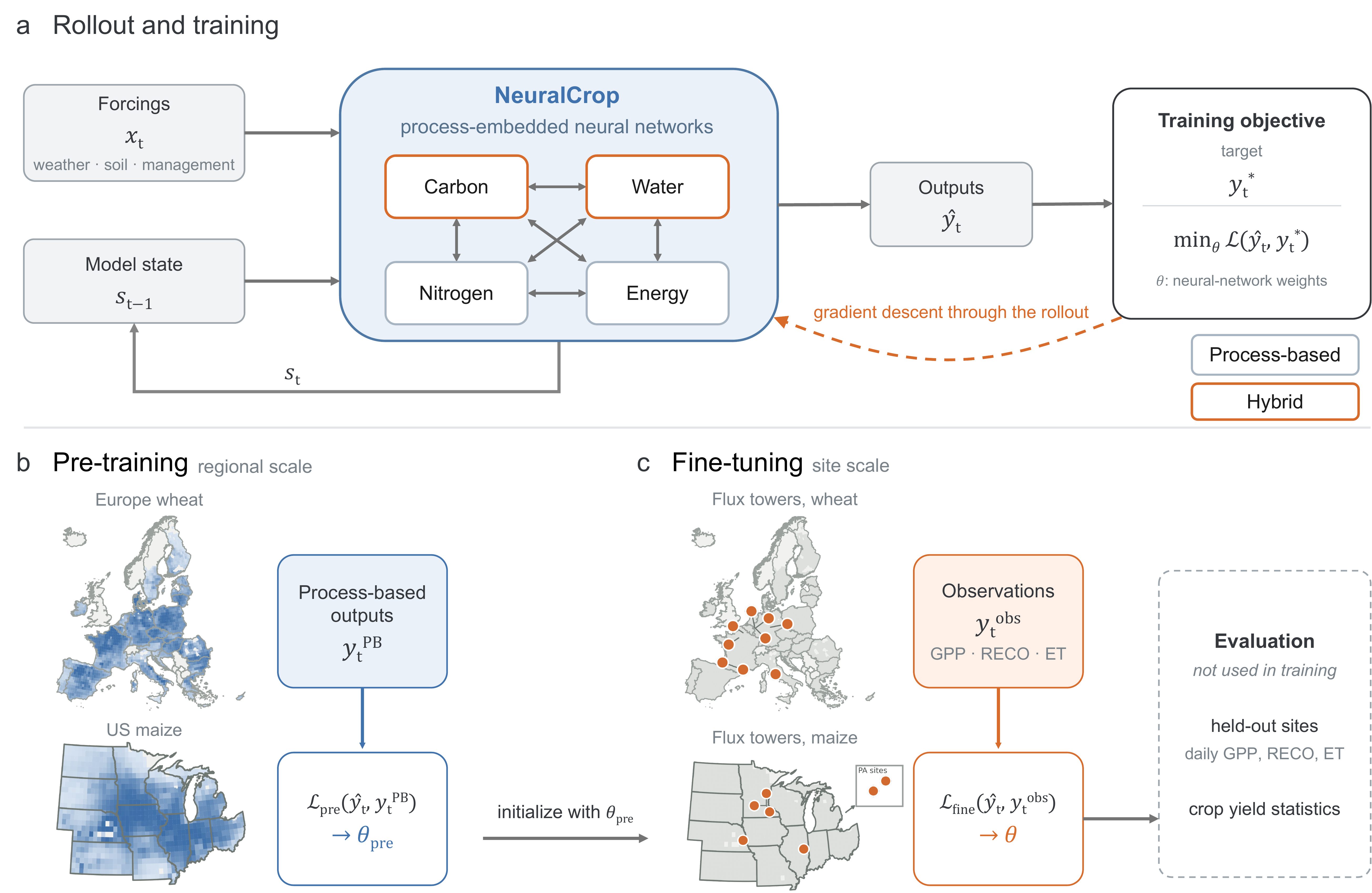}}
\caption{Schematic of the NeuralCrop rollout and two-stage training framework. \textbf{a,} At each time step $t$, NeuralCrop receives forcings $x_t$ and the previous model state $s_{t-1}$. The forcings include weather, soil properties, and field management, such as crop calendars, tillage, and fertilizer. Orange outlines identify the carbon and water processes that contain neural network components. Nitrogen and energy retain fully process-based representations. These coupled processes produce outputs $\hat{y}_t$ and an updated state $s_t$, which is carried forward to the next time step. Training minimizes $L(\hat{y}_t,y_t^*)$, where $y_t^*$ denotes the target for the corresponding training stage, and loss gradients propagate through the differentiable process calculations to update the neural-network parameters. \textbf{b,} During regional-scale pre-training, outputs generated by NeuralCrop in its fully process-based version provide the targets for $L_{\mathrm{pre}}$, yielding parameters $\theta_{\mathrm{pre}}$. \textbf{c,} During site-scale fine-tuning, the neural-network parameters are initialized with $\theta_{\mathrm{pre}}$ and further optimized against eddy-covariance observations of gross primary productivity (GPP), ecosystem respiration (RECO), and evapotranspiration (ET) using $L_{\mathrm{fine}}$. No simulated or observed crop yield data are used for training at any stage. The maps show nine wheat and seven maize flux-tower sites. The two Pennsylvania maize sites, which are included in the analysis but lie outside the main Corn Belt map, are shown in an inset. Held-out flux-tower observations and independent subnational yield statistics are used for evaluation.}
\label{hybrid_model_training_framework}
\end{figure}

NeuralCrop is built on the process representations of the LPJmL ("Lund-Potsdam-Jena managed Land") model. LPJmL explicitly simulates carbon, water, energy, and nitrogen flows for both natural vegetation and agricultural crops \cite{bondeau2007modelling, 2018LPJmL, von2018implementing, lutz2019simulating}. Its agricultural modules represent crop growth, management, and yield formation. LPJmL has been comprehensively evaluated at the global scale \cite{muller2014projecting, muller2017global, jagermeyr2017reconciling, schaphoff2018lpjml4, franke2020ggcmi, frieler2024scenario}. NeuralCrop augments these explicit process dynamics with trainable neural networks embedded in selected carbon and water processes. For example, neural networks represent the intercellular-to-atmospheric CO\textsubscript{2} concentration ratio ($\lambda$) and maximum Rubisco carboxylation capacity ($V_{\max}$) in photosynthesis, and provide corrections to crop respiration and transpiration. The relevant process-based components are implemented in a differentiable form to allow joint optimization within the simulation. The hybrid architecture and trainable components are described in detail in Supplementary Information sections~\ref{sec:hybrid_architecture} and~\ref{sec:components_of_NeuralCrop}.

Given limited high-quality agricultural observations, we adopt a two-stage strategy to train NeuralCrop, as illustrated in Fig.~\ref{hybrid_model_training_framework}. Pre-training NeuralCrop on simulations generated by its fully process-based version (which is essentially LPJmL) gives the trainable components a process-informed prior. This initializes the neural components to emulate the process-based calculations within the coupled model, allowing subsequent fine-tuning of these components using flux observation data. Online training jointly optimizes the learned components by propagating loss gradients through the relevant process calculations, accounting for their interactions within the model dynamics while retaining the process-based model structure~\cite{gelbrecht2023differentiable}. NeuralCrop supports hardware acceleration using Graphics Processing Units (GPU), which can massively speed up spatially distributed and ensemble simulations through efficient parallelization.

We train NeuralCrop to simulate wheat and maize at both site and regional scales. Training details are provided in Supplementary Information sections~\ref{sec:data}--\ref{sec:evaluation_metrics}. Pre-training results are presented in Supplementary Information Figs.~\ref{fig_EU_wheat_pretrain} and~\ref{fig_us_corn_pretrain}. We focus on the fine-tuned model below. At the site scale, we evaluate daily gross primary productivity (GPP), ecosystem respiration (RECO), and evapotranspiration (ET) using holdout sites grouped by geographic proximity, with all seasons at each held-out site excluded from training (Supplementary Information section~\ref{sec:site-level_validation}). At the regional scale, we evaluate wheat and maize yields against independent subnational statistics for European wheat regions and the US Corn Belt, and compare to process-based model benchmarks.

\section*{Results}

\subsection*{Independent site evaluation}

\begin{figure}[!t]
\centering
\makebox[\textwidth][c]{%
\includegraphics[width=1.2\textwidth]{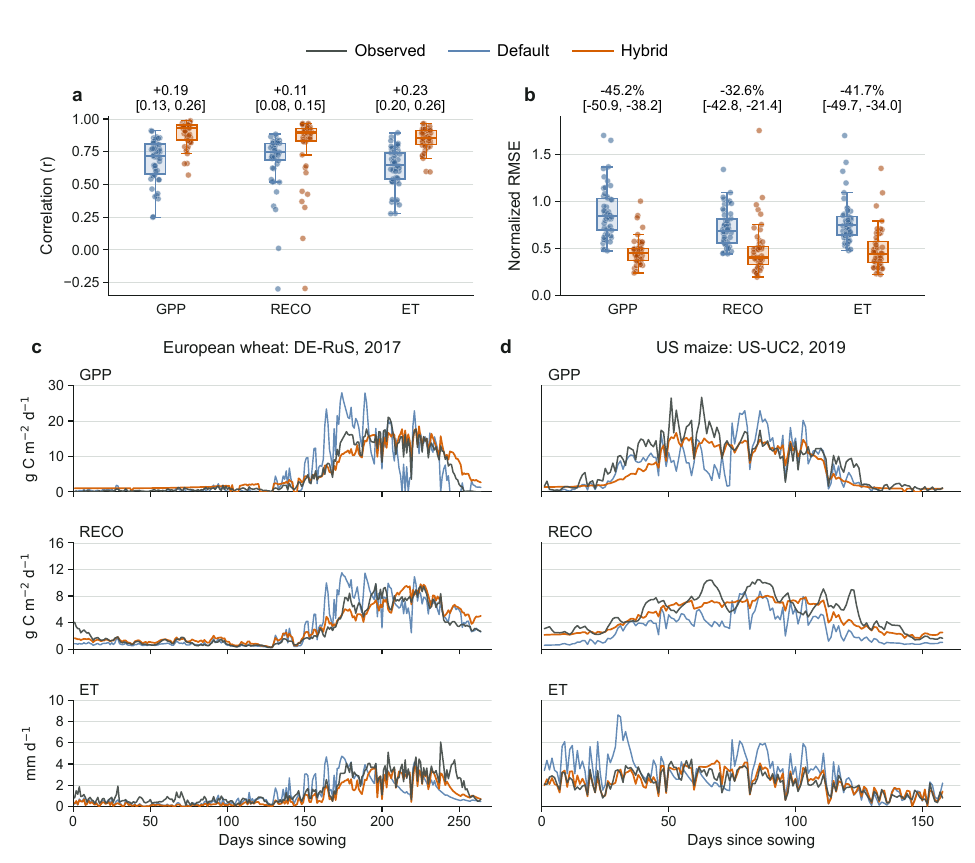}}
\caption{Held-out evaluation of daily carbon and water fluxes at flux-tower sites. \textbf{a,b}, Agreement between simulations and observations across 50 crop seasons held out from training. \textbf{a}, Pearson correlation, and \textbf{b}, normalized root mean square error (nRMSE) for gross primary productivity (GPP), ecosystem respiration (RECO), and evapotranspiration (ET). Each point in \textbf{a--b} represents one held-out season. Boxes show the median and interquartile range, and whiskers extend to 1.5 times the interquartile range. Annotations above \textbf{a} and \textbf{b} show the change in mean correlation and the percentage change in mean nRMSE, respectively, for the hybrid relative to the default simulations. Brackets show 95\% intervals from 10,000 bootstrap resamples of site groups. \textbf{c,d}, Examples of observed and simulated daily GPP, RECO, and ET during the held-out 2017 winter wheat season at DE-RuS in western Germany (\textbf{c}) and the 2019 maize season at US-UC2 in central Pennsylvania, USA (\textbf{d}). The default simulations use the fully process-based configuration, with all neural network components disabled. The hybrid model version reproduces the observations much more accurately than the default version.}
\label{fig:station_flux_validation}
\end{figure}

We first evaluate daily simulations of carbon and water fluxes using grouped site holdouts. The dataset comprises 50 crop seasons (29 wheat and 21 maize) assigned to 10 independent site groups (\hyperref[sec:held-out_evaluation]{Methods} and Supplementary Information section~\ref{sec:site-level_validation}). Across all held-out seasons, the hybrid simulations increase mean temporal correlation by 0.19 for GPP, 0.11 for RECO, and 0.23 for ET (Fig.~\ref{fig:station_flux_validation}a). The mean normalized RMSE is reduced by 45.2\%, 32.6\%, and 41.7\% respectively (Fig.~\ref{fig:station_flux_validation}b). Correlation improves in 48 of 50 seasons for GPP, 46 of 50 for RECO, and all 50 for ET, and the corresponding normalized RMSE decreases in 49, 46, and 50 of 50 seasons. As examples, Fig.~\ref{fig:station_flux_validation}c,d show one held-out wheat season and one held-out maize season selected to illustrate improvements across all three fluxes. The fully process-based (default) simulations show marked departures from observed GPP, RECO, and ET during the main growing period, including pronounced day-to-day fluctuations in GPP. The hybrid simulations more closely reproduce the observed trajectories of all three fluxes in both examples, although daily discrepancies remain.

To assess sensitivity to random training initialization, we repeat fine-tuning with two additional random seeds. Results from all three seeds are shown separately in Supplementary Information Fig.~\ref{fig:training_seed_sensitivity}, and the improvements remain consistent across seeds. Across all three seeds, we report the holdout results separately for wheat and maize (\hyperref[sec:held-out_evaluation]{Methods}). For wheat, the mean correlation gains are 0.180, 0.114, and 0.194 for GPP, RECO, and ET, with corresponding nRMSE reductions of 51.0\%, 42.4\%, and 32.1\%. For maize, the gains are 0.210, 0.120, and 0.266, with nRMSE reduced by 29.2\%, 18.0\%, and 55.9\%, respectively. The corresponding 95\% bootstrap intervals, obtained by resampling independent site groups, are reported in Supplementary Information Table~\ref{table:site_holdout_results}. For all three fluxes in both crops, the intervals show gains in correlation and reductions in nRMSE. These results indicate that the improvements are broadly distributed across the evaluated sites.

\subsection*{Regional yield evaluation}

\begin{figure}[h]
\centering
\makebox[\textwidth][c]{%
\includegraphics[width=1.2\textwidth]{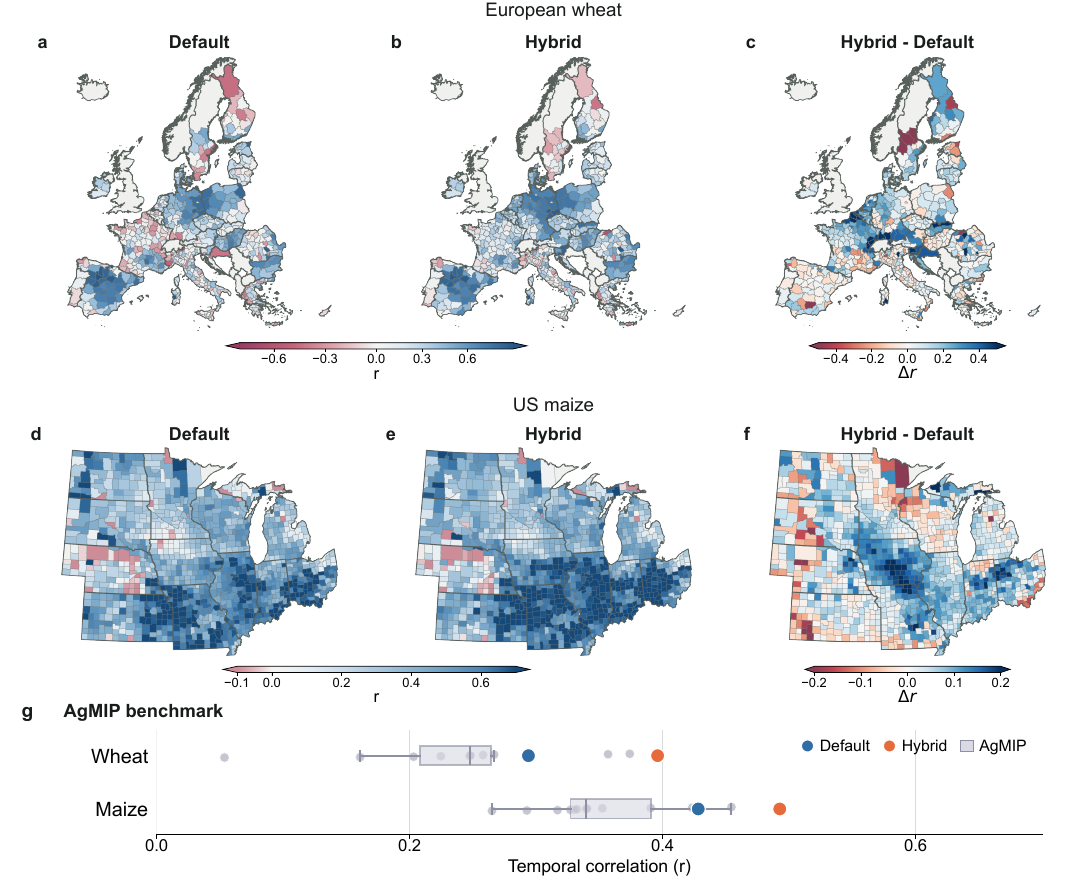}}
\caption{Evaluation of yield simulations and comparison with AgMIP models. \textbf{a,b}, Administrative-unit temporal correlation ($r$) between reported and simulated European wheat yields for the default and hybrid simulations over 1981--2019. \textbf{c}, Difference between the correlations obtained with the hybrid and fully process-based (default) model versions, respectively ($\Delta r$). Positive values indicate improvement by the hybrid version and occur across 71.4\% of mapped wheat area. \textbf{d,e}, Corresponding county-level default and hybrid simulations for US Corn Belt maize over 1981--2019. \textbf{f}, Corresponding $\Delta r$ map. Positive values occur across 84.0\% of mapped maize area. \textbf{g}, Crop area-weighted mean correlations for the default and hybrid simulations and the AgMIP GGCM distributions over the available 1981--2016 period. Points within the AgMIP boxplots denote individual GGCMs.}
\label{fig:regional_yield_skill}
\end{figure}

We fine-tune NeuralCrop using all available flux-tower seasons for the regional simulations. We then evaluate their ability to reproduce interannual yield variability against independent subnational yield data that are never used for training. The evaluation covers 530 European wheat reporting units defined according to the Nomenclature of Territorial Units for Statistics (NUTS)~\cite{ronchetti2024harmonized} and 1,047 US Corn Belt counties over 1981--2019. We additionally include AgMIP GGCMI Phase 3 simulations as an external model benchmark. Before comparison, simulated and reported yield series are linearly detrended separately to focus on interannual variability  (\hyperref[sec:yield_data_processing]{Methods}).

Although yield data are never used during training, the hybrid simulations show substantial spatial improvements over the default simulations (Fig.~\ref{fig:regional_yield_skill}). Temporal correlation is higher across 71.4\% of mapped European wheat area and 84.0\% of mapped US maize area (Fig.~\ref{fig:regional_yield_skill}c,f). The crop area-weighted mean correlation increases from 0.303 to 0.392 for wheat ($\Delta r=+0.089$, 95\% CI [0.034, 0.137]) and from 0.415 to 0.479 for maize ($\Delta r=+0.064$, 95\% CI [0.033, 0.089]). The nRMSE difference maps show a broadly consistent pattern of error reduction for both crops (Supplementary Information Fig.~\ref{fig:regional_yield_nrmse}). A distance analysis shows that positive correlation gains remain even among administrative units farthest from the crop-specific training sites (Supplementary Information Fig.~\ref{fig:yield_site_distance}). These gains show that fine-tuning against observed daily carbon and water fluxes can improve the simulation of interannual yield variability, although grain yield depends on the distribution of assimilated carbon. 

We also compare our hybrid and default simulations with state-of-the-art GGCMs taking part in AgMIP GGCMI Phase~3 \cite{jagermeyr2021climate}. NeuralCrop shows highly competitive performance in crop area-weighted temporal correlation and nRMSE for both European wheat and US maize (Fig.~\ref{fig:regional_yield_skill}g and Supplementary Information Fig.~\ref{fig:regional_yield_nrmse}g). Specifically, NeuralCrop increases area-weighted mean correlation by 59.8\% with respect to the AgMIP median for wheat and 45.1\% for maize, and ranks first in the AgMIP models for both crops. For comparison, the fully process-based version increases 18.6\% for wheat and 26.1\% maize. Fine-tuning NeuralCrop on fluxnet observations thus substantially improves overall yield simulations for wheat and maize. Across three random seeds, the distributions of administrative-unit correlation and nRMSE are also consistent for both crops (Supplementary Information Fig.~\ref{fig:yield_seed_sensitivity}). 

\subsection*{Component sensitivity analysis}

\begin{figure}[!t]
\centering
\makebox[\textwidth][c]{%
\includegraphics[width=1.2\textwidth]{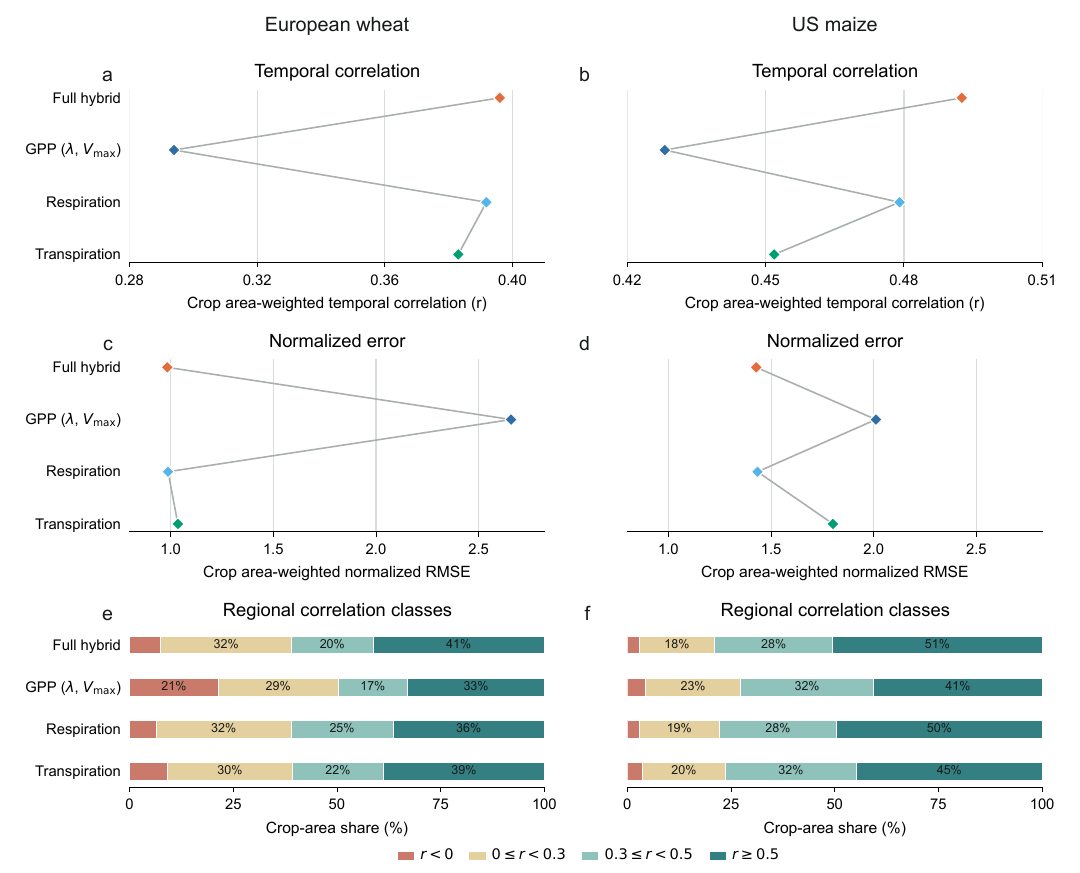}}
\caption{Sensitivity of regional yield skill to the individual neural network components. The left column shows European wheat and the right column US maize. \textbf{a,b}, Crop area-weighted temporal correlation between detrended simulated and reported annual yields for the full hybrid version and three variants evaluated during inference. In the GPP ($\lambda$, $V_{\max}$) variant, process-based calculations replace the neural network predictions of these variables. In the respiration and transpiration variants, the neural corrections to crop respiration and to root-water supply in crop transpiration, respectively, are removed. \textbf{c,d}, Corresponding crop area-weighted normalized RMSE (nRMSE). Higher correlation and lower nRMSE indicate better agreement. \textbf{e,f}, Percentage of cultivated area in four administrative-unit temporal-correlation classes ($r<0$, $0\leq r<0.3$, $0.3\leq r<0.5$, and $r\geq0.5$) for the four versions.}
\label{fig:component_yield_magnitude}
\end{figure}

We assess the contribution of each learned neural network component of NeuralCrop to regional yield skill across four configurations, including the full hybrid version with all three evaluated neural components active, and three variants in which one learned component at a time reverted its process-based calculation (Fig.~\ref{fig:component_yield_magnitude}a--d). The GPP component is comprised of multilayer perceptrons (MLPs) for $\lambda$ and $V_{\max}$, and the other two components apply neural corrections to crop respiration and to the root-water supply in crop transpiration, respectively. The reported mean correlation is the crop area-weighted average of temporal correlations between detrended simulated and reported annual yields calculated separately within each administrative unit. Reverting $\lambda$ and $V_{\max}$ to their process-based calculations causes the largest decrease in the mean correlation and increase in nRMSE relative to the full hybrid model for both crops. Removing the transpiration correction also reduces the skill, particularly for US maize, whereas removing the respiration correction has a smaller effect.

The correlation classes show a similar pattern across cultivated areas (Fig.~\ref{fig:component_yield_magnitude}e,f). Relative to the full hybrid model, reverting the GPP component to its process-based calculations reduces the share of wheat and maize area with $r\geq0.5$ and shifts more area into lower-correlation classes. The increase in negative-correlation area is particularly apparent for wheat. Removing the transpiration correction also shifts maize area toward lower-correlation classes, while removing the respiration correction leaves results comparatively close to the full hybrid model for both crops. These results show that regional yield skill is most sensitive to GPP in NeuralCrop, with a smaller sensitivity to the transpiration correction, particularly for maize. In a separate analysis, changes in the correlations of simulated seasonal GPP, crop respiration, and crop transpiration with reported yield are positively associated with changes in simulated yield correlation across administrative units (Supplementary Information Fig.~\ref{fig:process_yield_transfer}).

\clearpage
\subsection*{Yield skill across moisture conditions}

\begin{figure}[!htp]
\centering
\makebox[\textwidth][c]{%
\includegraphics[width=1.03\textwidth]{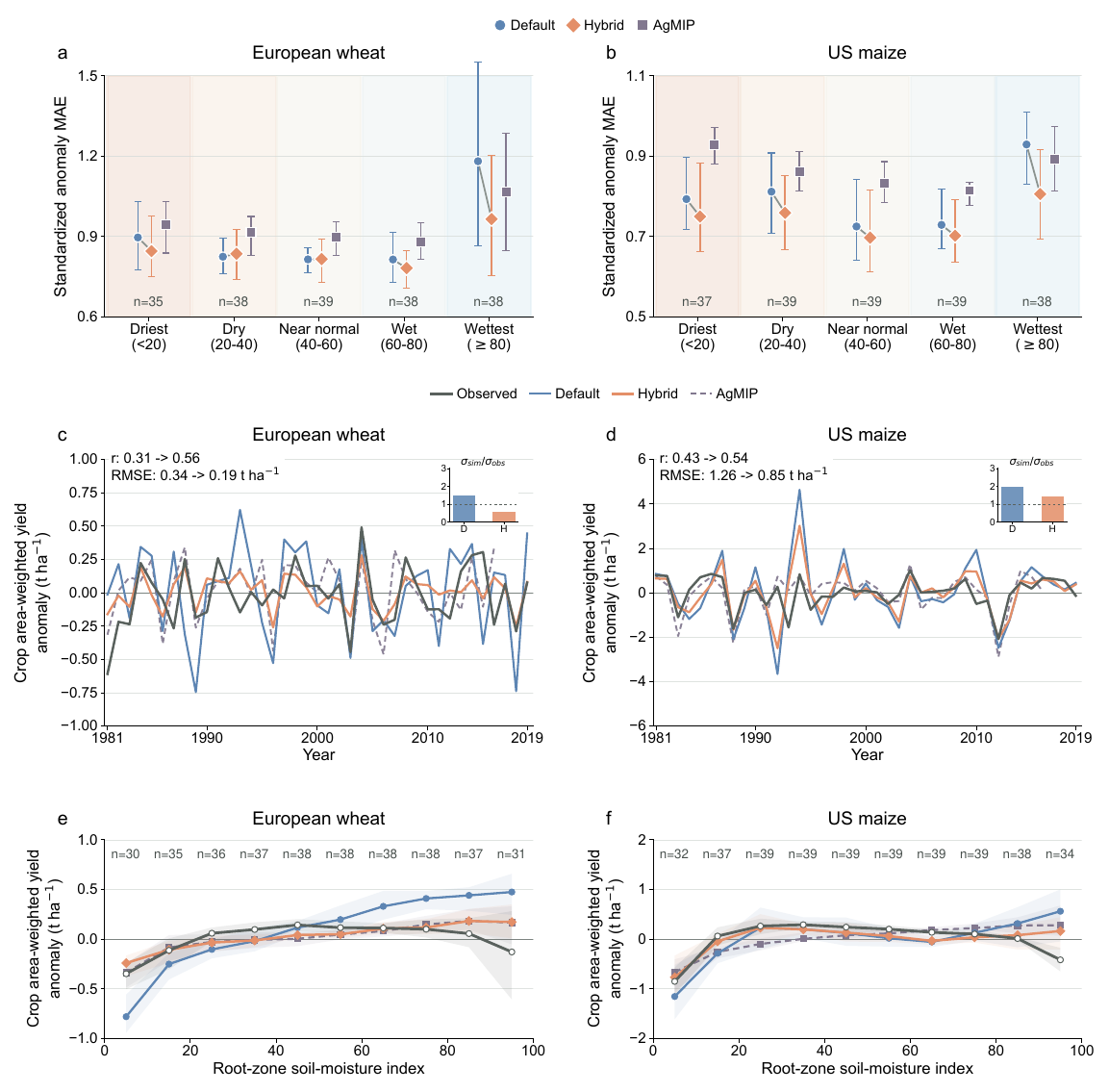}}
\caption{Yield anomalies across soil moisture conditions. \textbf{a,b}, Crop area-weighted mean absolute error (MAE) of standardized yield anomalies for European wheat and US maize, grouped by the growing season ERA5-Land root zone soil-moisture index of each administrative-unit year. The index is the crop area-weighted mean of grid-cell soil-moisture percentile ranks relative to 1981--2010. Anomalies are standardized separately for each yield series within each administrative unit. Points show the estimates and bars show 95\% intervals from 5,000 country- or state-cluster bootstrap resamples. Default and hybrid simulations are evaluated over 1981--2019, whereas AgMIP ensemble median is evaluated over 1981--2016. \textbf{c,d}, Crop area-weighted regional mean yield anomalies from reported statistics, default, hybrid, and AgMIP ensemble median. Insets show the ratio of simulated to observed interannual standard deviation. Annotations in the top left of each panel give the temporal correlation ($r$) and RMSE between the reported regional mean yield time series and the corresponding series from the default and hybrid simulations over 1981--2019 (default $\rightarrow$ hybrid). For European wheat, correlation (RMSE) is thus improved by 80.6 (44.1)\%, while for US maize correlation (RMSE) is improved by 25.6 (32.5)\% by NeuralCrop compared to the fully process-based baseline.
\textbf{e,f}, Crop area-weighted mean yield anomaly in 10-point soil-moisture index bins. Shading shows 95\% country- or state-cluster bootstrap intervals. The $n$ labels in \textbf{a,b,e,f} give the number of distinct calendar years contributing at least one administrative-unit record to each moisture bin in the evaluation over 1981--2019.}
\label{fig:moisture_extremes}
\end{figure}

The yield prediction performance of NeuralCrop is further evaluated across different soil-moisture conditions. For each grid cell, we rank the growing-season ERA5-Land root-zone soil moisture in each evaluated year (1981--2019) against the corresponding 1981--2010 values and express the rank as a percentile \cite{munoz2021era5}. We then average these grid-cell percentile ranks within each administrative unit using crop-area weights to obtain a relative soil-moisture index on a 0--100 scale. Each unit-year is assigned to one of five equal-width classes of this index. Within each class, we calculate the crop area-weighted MAE between standardized simulated and reported yield anomalies (Fig.~\ref{fig:moisture_extremes}a,b). The default, fully process-based simulations show a broadly U-shaped error pattern for both crops, with relatively low standardized-anomaly MAE near normal conditions and overall higher errors as soil moisture shifts toward either extreme (Fig.~\ref{fig:moisture_extremes}a,b). The AgMIP ensemble median shows a comparable pattern. The hybrid NeuralCrop simulations have lower MAE point estimates than the default and AgMIP simulations across conditions but especially at both extremes in the observed 1981–2019 records, especially under the wettest conditions.

The regional crop area-weighted anomaly records provide temporal context beyond the moisture-class averages (Fig.~\ref{fig:moisture_extremes}c,d). For wheat, correlation increased from 0.31 to 0.56 and RMSE decreased from 0.34 to 0.19~t~ha$^{-1}$ for the 1981--2019 period. For maize correlation increased from 0.43 to 0.54 and RMSE decreased from 1.26 to 0.85~t~ha$^{-1}$ for the 1981--2019 period. The baseline anomaly standard deviation was 1.51 and 2.0 times the observed value for wheat and maize, respectively, compared with 0.58 and 1.42 for the hybrid simulation. The hybrid simulations therefore reduce the excessive year-to-year variability of the default simulations, although they partially underestimate the amplitude of wheat anomalies. The lower wheat RMSE hence should be considered alongside the correlation gain.

The response curves show where simulated yield anomalies depart from observations (Fig.~\ref{fig:moisture_extremes}e,f). The observed point estimates are negative at both ends of the soil-moisture range for both crops, although the wettest wheat estimate has wide uncertainty. The default simulations show a much steeper increase in yield anomalies from dry to wet conditions than is observed, whereas the hybrid simulations have a flatter response in accordance with observations. This contrast persists when anomalies are expressed as percentages of trend yield (Supplementary Information Fig.~\ref{fig:moisture_response_percent}). At the dry end, the hybrid estimates are closer to observations for both crops, nearly matching the observed maize anomaly ($-0.77$ versus $-0.85$~t~ha$^{-1}$). At the wet end, however, the default, hybrid, and AgMIP ensemble estimates remain positive despite negative observed anomalies for both crops. The $n$ labels in Fig.~\ref{fig:moisture_extremes}a,b,e,f show the number of calendar years represented in each moisture bin. The driest and wettest 10-point index bins contain records from 30--34 years, but crop-area contributions vary considerably among those years (Supplementary Information Table~\ref{table:moisture_year_support}).

\clearpage
\section*{Discussion}

NeuralCrop combines coupled process-based representations of carbon, water, nitrogen, and energy adapted from LPJmL \cite{bondeau2007modelling,von2018implementing} with neural network components representing GPP, respiration, and transpiration, trained on process-based simulations and physiological variables from flux-tower observations. The neural networks are trained and act within selected daily process calculations, updating model states that are propagated forward through the growing season. Pre-training on outputs from the default fully process-based model version provides an initialization before fine-tuning with flux-tower observations. In grouped site holdouts, the hybrid simulations show higher temporal correlations and lower nRMSEs for daily GPP, RECO, and ET than the fully process-based simulations in most evaluated seasons (Fig.~\ref{fig:station_flux_validation}a,b). The improvements remain consistent across three training seeds for both wheat and maize.

Interannual yield variability is relevant to food security under climate change \cite{proctor2025climate}. Although reported yield data are not used for training, the regional hybrid simulations show closer agreement with independent records of interannual wheat and maize yield variability. Temporal yield correlation is higher across 71.4\% of mapped European wheat area and 84.0\% of mapped US maize area (Fig.~\ref{fig:regional_yield_skill}c,f). Regional yield-skill distributions are similar across three training seeds. For both crops, crop area-weighted mean correlation gains remain positive even in the quarter of mapped crop area farthest from the crop-specific flux-tower sites (Supplementary Information Fig.~\ref{fig:yield_site_distance}). Together, these results suggest that fine-tuning against physiological observations from flux towers can improve the model's ability to capture observed interannual yield variability in the evaluated regions.

Regional yield performance also varies with soil moisture in the root zone during the growing season (Fig.~\ref{fig:moisture_extremes}). For both crops, the default simulations and the AgMIP ensemble median show broadly higher MAE of standardized yield anomalies toward the dry and wet ends of the soil moisture range. The hybrid simulations follow a similar pattern but have lower MAE point estimates at both ends, particularly in the wettest class. The response curves show that, at the dry end, the hybrid NeuralCrop simulations more closely reproduce the negative yield anomalies observed for both crops. At the wet end, however, the default, hybrid, and AgMIP ensemble median estimates of yield anomalies remain positive while the observed point estimates are negative for both crops (Fig.~\ref{fig:moisture_extremes}e,f). This disagreement between simulated and observed yield anomalies highlights a remaining limitation in simulating yield responses under high soil moisture. These results characterize performance across soil-moisture conditions during 1981--2019, but do not establish how the model would perform during extreme events outside this evaluation period.

Our results also reveal important limitations. Although the hybrid NeuralCrop simulations increase the temporal correlation of regional interannual yield anomalies from 0.31 to 0.56 for wheat and from 0.43 to 0.54 for maize (Fig.~\ref{fig:moisture_extremes}c,d), discrepancies between simulated and observed yields remain. %
Yield formation depends on the allocation of assimilated carbon among plant organs, a process that remains largely empirically parameterized in many GGCMs. Direct allocation observations are scarce, making the link between photosynthesis and yield difficult to constrain. Phenology and responses to water conditions add further uncertainty. The component experiments indicate that regional yield skill is most sensitive to the data-driven GPP component in NeuralCrop (Fig.~\ref{fig:component_yield_magnitude}), but they do not identify the processes responsible for the remaining yield errors. Although the hybrid simulations have lower MAE point estimates in the wetter classes, they still miss the observed negative yield anomalies under the wettest conditions. The limited representation of excess-water stress in crop models may contribute to the errors under high soil moisture \cite{garcia2025gaps}.

Addressing these limitations will require more direct field measurements of crop growth and stress responses, alongside further development of process representations. NeuralCrop can readily incorporate new process knowledge, additional trainable components, and new observational information as suitable data become available. NeuralCrop's implementation also fully supports GPU acceleration, making computationally demanding tasks such as large ensemble simulations more feasible. For example, GPU-accelerated NeuralCrop ran nearly two orders of magnitude faster than CPU-based LPJmL for the same large-scale 20-year daily simulations (Supplementary Information Fig.~\ref{fig_time_efficiency}). Given the limited availability of high-quality crop flux observations, fine-tuning in this study relied on data from only 16 flux-tower sites. We believe NeuralCrop could substantially further improve crop simulations and support food risk assessments under climate change as additional observations become available.

\section*{Methods}

\subsection*{NeuralCrop architecture}

We develop NeuralCrop by first reimplementing the crop and soil processes of LPJmL ("Lund--Potsdam--Jena managed Land")~\cite{bondeau2007modelling, 2018LPJmL, von2018implementing, lutz2019simulating} from C into Julia \cite{bezanson2017julia}. We then reformulate selected modules to support automatic differentiation (AD) through the daily model dynamics. The resulting implementation is therefore derived from LPJmL, with adaptations for differentiation (Supplementary Information Section~\ref{sec:components_of_NeuralCrop}). 

We then implemented neural networks in the model that represent very specific individual processes, and operate within daily process calculations. This makes NeuralCrop a hybrid model. For example, the GPP component contains separate fully connected MLPs for $\lambda$ and $V_{\max}$, while bounded neural corrections modify crop respiration and transpiration. Each individual neural network used in the evaluated hybrid version has two hidden layers with 64 neurons per layer. At each daily time step, outputs from active neural networks influence updates to prognostic model states, affecting subsequent crop growth and yield formation.

NeuralCrop uses Zygote.jl \cite{Zygote.jl-2018} for reverse-mode automatic AD through the differentiable model trajectory. During the forward rollout, process calculations and active neural networks update prognostic states from one day to the next. AD then propagates the loss gradient backward through these calculations to update the neural-network parameters, accounting for their effects on simulated states. The implementation also supports graphics-processing-unit (GPU) execution.

\subsection*{Two-stage training approach}

Due to the limited high-quality observational data, we adopt a two-stage training approach (Fig.~\ref{hybrid_model_training_framework}). During the pre-training stage, we train NeuralCrop against targets generated by its fully process-based version by minimizing a loss function composed of two terms (Supplementary Information~\ref{sec:loss_function}, Eq.~\ref{eq:loss1}). The first term captures discrepancies between the hybrid and process-based outputs, while the second term captures deviations from the mass balance of crop carbon. We then fine-tune selected neural components against eddy-covariance observations of daily gross primary productivity (GPP), ecosystem respiration (RECO), and evapotranspiration (ET) (Supplementary Information~\ref{sec:loss_function}, Eq.~\ref{eq:loss2}). The processing of flux data is described in detail in Supplementary Information~\ref{sec:data_pre-processing}. All loss terms are based on mean squared error (MSE) (Supplementary Information~\ref{sec:loss_function}, Eq.~\ref{eq:mse}).

For pre-training, we use 365-day rollouts beginning at sowing, allowing each rollout to span the crop season. Thereafter, site-level fine-tuning uses season-specific windows defined by sowing and harvest events, with the loss evaluated only on days with valid flux observations. Although shorter rollouts reduce computation, they produced greater fluctuations in training and validation loss. The 365-day rollouts showed comparably stable convergence, supporting our choice of 365 days for pre-training (Supplementary Information~\ref{sec:training_rollouts}).

\subsection*{Pre-training data generation}

To generate the data used for pre-training, we run NeuralCrop in its fully process-based mode, with all neural network components disabled. In this mode, NeuralCrop uses the crop and soil process formulations of LPJmL~\cite{bondeau2007modelling,2018LPJmL,von2018implementing,lutz2019simulating}, reimplemented in Julia to support training of the embedded neural networks. Driven by climate forcing, soil properties, land use, and field management (e.g., crop calendar, tillage, residue, and fertilizer), it produces daily trajectories of photosynthesis, carbon allocation, evapotranspiration, soil moisture, crop growth, and yield formation.

LPJmL is used to provide initial states for the NeuralCrop simulations. We ran LPJmL version 5 globally at \(0.5^\circ \times 0.5^\circ\) spatial resolution with a daily time step \cite{von2018implementing}. An initial 5000-year spin-up under potential natural vegetation (i.e., without land use) brought the soil carbon, nitrogen, and water pools into dynamic equilibrium. Starting from these states, a second 390-year spin-up introduced the effects of historical land-use change on the pools. Further details are provided in Section~\ref{sec:data} of the Supplementary Information.

\subsection*{Held-out evaluation}\label{sec:held-out_evaluation}

We evaluate daily GPP, RECO, and ET for 50 crop seasons (29 wheat and 21 maize) from 16 flux-tower sites. Flux towers located close to one another and experiencing broadly similar climatic conditions are assigned to the same group to prevent neighboring sites from appearing in both training and evaluation. This yields 10 site groups, including six for wheat and four for maize. We use leave-one-group-out evaluation. In each fold, all seasons from the held-out group are excluded from fine-tuning. Each fold is fine-tuned for 150 epochs with three random seeds, and metrics are calculated separately for each held-out season and seed.

We further calculate mean correlation and nRMSE for each physical site across its held-out seasons and three training seeds. For a group containing multiple sites, we average their correlation and nRMSE values to obtain one result for that group. We obtain 95\% bootstrap intervals for the differences between the default and hybrid simulations by resampling whole site groups with replacement 10,000 times separately for wheat and maize. Site identities and group assignments are listed in Supplementary Information Table~\ref{table:station_groups}.

\subsection*{Regional yield evaluation}

We evaluate regional yield simulations against subnational reported yields for European wheat and US Corn Belt maize. The harmonized European crop statistics~\cite{ronchetti2024harmonized} and USDA-NASS county records are described in Supplementary Information Sections~\ref{sec:subnational_data} and~\ref{sec:county-level_data}. Reported yields are not used for training but serve as observational benchmarks. However, they remain subject to uncertainties in reporting, administrative boundary definitions, and spatial aggregation. The process-based and hybrid simulations are evaluated over 1981--2019. We also compare these simulations with the AgMIP ensemble. AgMIP includes 13 GGCMs, i.e., ACEA \cite{mialyk2024water}, CROVER~\cite{okada2015modeling}, EPIC-IIASA \cite{balkovivc2014global}, ISAM~\cite{song2013implementation}, LDNDC~\cite{haas2013landscapedndc}, LPJmL~\cite{von2018implementing}, LPJ-GUESS~\cite{ma2022modeling}, PEPIC \cite{williams1989epic}, DSSAT-Pythia \cite{joshi2026pythia}, pDSSAT~\cite{jones2003dssat}, PROMET~\cite{mauser2015global}, CYGMA1p74 \cite{iizumi2017responses} and SIMPLACE-LINTUL5~\cite{enders2023simplace}. For the component sensitivity analysis, we evaluate the trained full hybrid configuration and three variants in which the GPP, crop respiration, or transpiration component is disabled during inference. All four configurations are evaluated against the same administrative-unit yield records over 1981--2019 for each crop.

\subsection*{Regional yield processing}\label{sec:yield_data_processing}

We aggregate gridded yields to the administrative units of the reported statistics using 2015 harvested area weights, accounting for the exact overlap between grid cells and administrative boundaries. Grid cells with less than 10 ha of cultivated area are excluded before aggregation. Rainfed and irrigated yields are combined before aggregation. Simulated and reported series are paired over common valid years without interpolation. To quantify interannual variability, we detrend the simulated and reported yield series separately using linear trends. Administrative units with fewer than five paired years are excluded. Yield anomalies are expressed as detrended departures in t~ha$^{-1}$. We obtain 95\% spatial-cluster bootstrap intervals for the regional mean change by resampling whole countries for European wheat ($n=26$) or whole states for US maize ($n=12$) with replacement 5,000 times. In each resample, we retain all administrative units in the selected countries or states and recalculate the crop area-weighted mean change in administrative-unit temporal correlation.

We take soil moisture of the root zone during the growing season from ERA5-land \cite{munoz2021era5}, convert it to an empirical percentile relative to 1981--2010 at each $0.5^\circ$ grid cell, and then average these percentiles over administrative units using crop-area weights (Supplementary Information Section~\ref{sec:soil_moisture_classes}). We use the crop area-weighted mean of these grid-cell percentile ranks as a relative soil-moisture index on a 0--100 scale for each administrative unit. For the moisture-class analysis, detrended reported and simulated yield anomalies are standardized separately by the standard deviation of their respective series within each unit. Administrative-unit years are grouped into five 20-point index classes, and the spatial mean absolute error (weighted by crop area size) is calculated for each class. For the response curves, administrative-unit years are instead grouped into ten 10-point index bins, and yield anomalies are averaged using the same area weights. For each moisture class or bin, we count the distinct calendar years contributing at least one administrative-unit record and calculate each year's share of the total crop-area weight within that class or bin. We apply the same country- or state-level spatial-cluster bootstrap procedure to obtain 95\% intervals for both moisture analyses.

\subsection*{Evaluation metrics}

For daily site fluxes, we report Pearson correlation and normalized RMSE (nRMSE) without detrending. For regional interannual yield variability, we also calculate Pearson correlation and nRMSE after detrending. Administrative-unit metrics are summarized using 2015 crop-area weights. Definitions and uncertainty procedures are given in Supplementary Information section~\ref{sec:evaluation_metrics}.

\section*{Data availability}

The climate forcing (GSWP3--W5E5) and crop management inputs were obtained from ISIMIP3a (\url{https://www.isimip.org/gettingstarted/input-data-bias-adjustment/}). The ERA5-Land monthly averaged soil-moisture data were downloaded from the Copernicus Climate Data Store (\url{https://cds.climate.copernicus.eu/datasets/reanalysis-era5-land-monthly-means}).

The daily eddy-covariance flux and meteorological data used to fine-tune NeuralCrop were obtained from FLUXNET products through the FLUXNET Shuttle (\url{https://data.fluxnet.org/}).

The European subnational wheat statistics were downloaded from \url{https://data.jrc.ec.europa.eu/dataset/685949ff-56de-4646-a8df-844b5bb5f835}, which provides annual wheat statistics on area, production, and average yield across three levels of the Nomenclature of Territorial Units for Statistics (NUTS). The subnational corn statistics of the US Corn Belt were obtained from \url{https://quickstats.nass.usda.gov/}. The AgMIP GGCMI Phase 3 yield simulations used for comparison were obtained from ISIMIP3 (\url{https://www.isimip.org/outputdata/}).

\section*{Code availability}

The NeuralCrop Julia source code is publicly available at \url{https://github.com/yunan-l/NeuralCrop.jl}. The code used to generate main figures, is archived at Zenodo \url{https://doi.org/10.5281/zenodo.23823351}.

\section*{Supplementary information}

The manuscript is accompanied by Supplementary Information that describes in detail how to develop, train, and validate NeuralCrop, as well as additional results.

\section*{Author contribution}

Y.L., S.B., M.B., M.G., P.H. and N.B. conceived and designed the study. Y.L. wrote the model code, processed the data, trained the models and analysed the results. J.H. provided the yield comparison code. All authors discussed the results. Y.L. wrote the manuscript with contributions from all authors.

\section*{Acknowledgements}

Y.L. acknowledges funding from the program of the China Scholarship Council (no.202303250017). N.B. and S.B. acknowledge funding from the Volkswagen Foundation. This is ClimTip contribution \#X; the ClimTip project has received funding from the European Union’s Horizon Europe research and innovation programme under grant agreement no.101137601. The authors gratefully acknowledge the Ministry of Research, Science and Culture (MWFK) of Land Brandenburg for supporting this project by providing resources on the high performance computer system at the Potsdam Institute for Climate Impact Research.

\clearpage
\bibliography{sn-bibliography}%

\backmatter
\clearpage

\appendix

\begin{bibunit}[unsrt]

\section*{\centering Supplementary information}
\renewcommand{\thefigure}{S\arabic{figure}}
\renewcommand{\thetable}{S\arabic{table}}
\renewcommand{\theequation}{S\arabic{equation}}

\renewcommand{\theHfigure}{supp.figure.\arabic{figure}}
\renewcommand{\theHtable}{supp.table.\arabic{table}}
\renewcommand{\theHequation}{supp.equation.\arabic{equation}}
\setcounter{figure}{0}

\addtocontents{toc}{\protect\setcounter{tocdepth}{3}}
\renewcommand{\contentsname}{}
\tableofcontents

\newpage

\section{Hybrid architecture} \label{sec:hybrid_architecture}

\subsection{Dynamical Core of NeuralCrop}

NeuralCrop distinguishes between diagnostic and prognostic variables. Diagnostic variables, such as leaf area index (LAI) and gross primary productivity (GPP), represent instantaneous system states that are not explicitly dependent on the system's temporal evolution. Prognostic variables, such as carbon pools, nitrogen pools and soil water pools, govern the dynamic evolution of the system states over time. They are represented by coupled ordinary differential equations (ODEs) that form the dynamical core of NeuralCrop in crop carbon, nitrogen, and water cycles, as described in detail in Section \ref{sec:components_of_NeuralCrop}. At each time step, diagnostic variables are updated based on the computed prognostic variables.

\subsection{Neural network}

NeuralCrop supports fully connected neural networks and neural ordinary differential equations (NODEs) \cite{chen2018neural_s} to replace or augment selected process representations. Each individual neural network in the implementation has two hidden layers with 64 neurons per layer. To solve the NODEs, we use a forward Euler solver with a fixed time step $\Delta t = 1$ day. At each time step $t$, the prognostic variable $\mathbf{X}(t)$ is updated according to:
\begin{align}
\mathbf{X}(t + \Delta t) = \mathbf{X}(t) + \Delta t \cdot f_\theta(\mathbf{X}(t), t)
\end{align}
where $f_\theta$ denotes the MLP that outputs the learned tendencies. The individual architectures and their coupling to the process dynamics are described below.

\subsection{Implementation}

We implement the ML components of NeuralCrop using Lux.jl \cite{pal2023lux, pal2023efficient}, a neural network framework built entirely with pure functions that are friendly to both compilers and automatic differentiation. We train the ML components of NeuralCrop using Enzyme.jl~\cite{moses2021enzyme_s}, a high-performance automatic differentiation (AD) framework in Julia that enables efficient gradient-based optimization, accelerated by CUDA.jl \cite{besard2018juliagpu, besard2019prototyping}. During training, the gradients of neural networks are computed via direct reverse-mode differentiation, including the forward Euler solver used for the NODEs (i.e., gradients are obtained by backpropagating through the Euler solver rather than using adjoint methods). To support full graphics-processing-unit (GPU) compatibility and improve computational efficiency, we implement the process-based components using KernelAbstractions.jl~\cite{Churavy_KernelAbstractions_jl}, which allows both neural networks and process-based parts to be parallelized and executed within a unified GPU-accelerated framework.

\clearpage
\section{Trainable components of NeuralCrop} \label{sec:components_of_NeuralCrop}

\begin{figure}[H]
\centering
\includegraphics[width=0.95\textwidth]{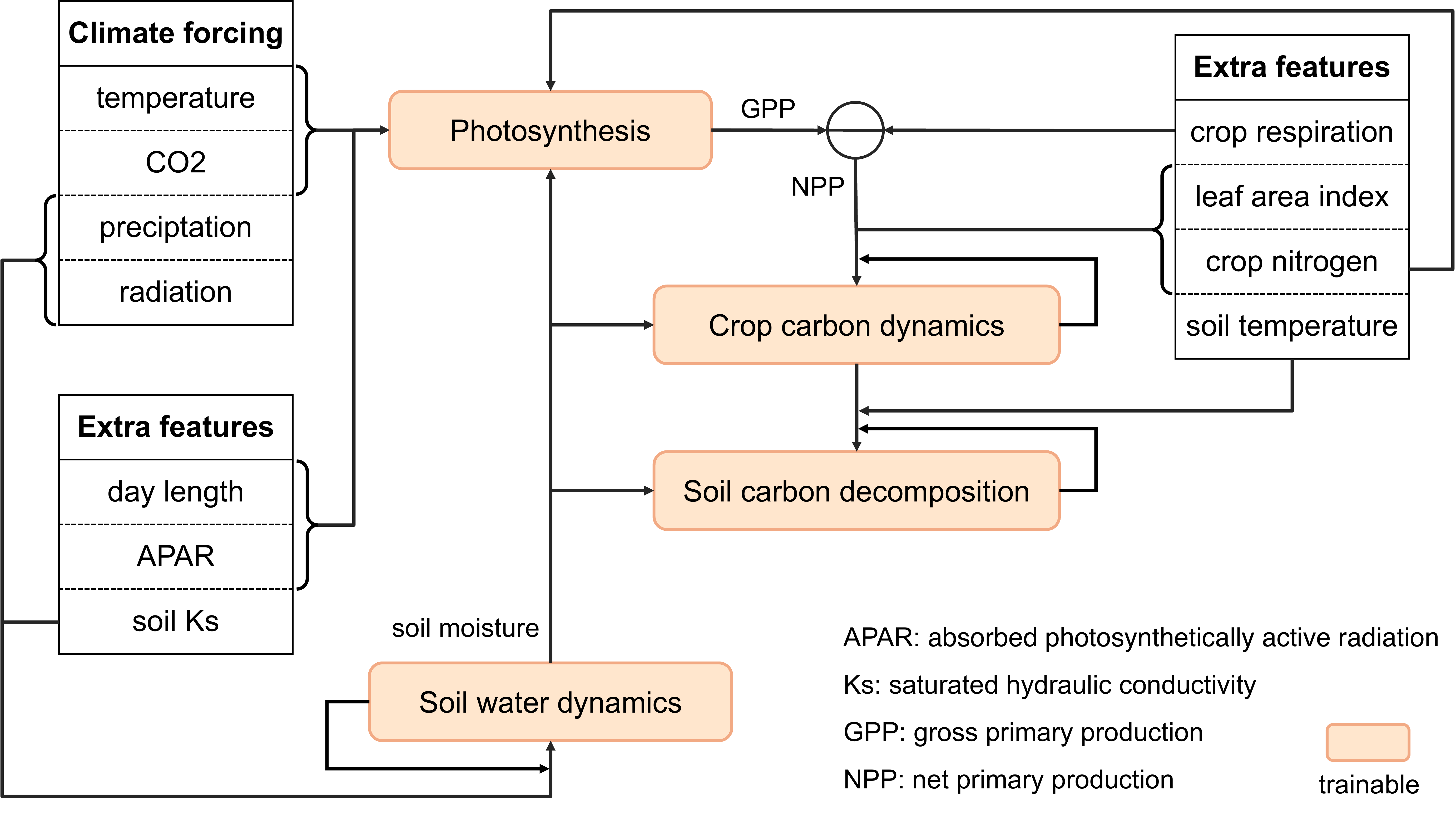}
\caption{Schematic of data flow through the trainable components of NeuralCrop. The schematic also includes optional components that are not activated in the main-text simulations.}
\label{fig_hybrid_structure}
\end{figure}

A detailed description of the model components of LPJmL can be found in Schaphoff et al.~\cite{schaphoff2018lpjml4_s}. In this section, we focus on the components replaced or augmented by neural networks within NeuralCrop. The overall data flow in the learned components of NeuralCrop is shown in Fig.~\ref{fig_hybrid_structure}. In the simulations presented in the main text, we activate only the neural components for GPP, crop respiration, and crop transpiration, because the available observations provide targets for fine-tuning them. The other neural representations described below remain inactive in these simulations.

\subsection{Photosynthesis} \label{sec:photosynthesis}
The photosynthesis scheme follows the ``big leaf'' formulation developed by Farquhar et al.~\cite{farquhar1980biochemical, farquhar1982modelling} and extended by Collatz et al.~\cite{collatz1991physiological, collatz1992coupled} and Prentice et al.~\cite{haxeltine1996general, prentice2000carbon}. Daily gross photosynthesis, $A_{\mathrm{gd}}$, is jointly limited by the light limited rate, $J_{\mathrm{E}}$, and the Rubisco limited rate, $J_{\mathrm{C}}$:
\begin{align}
J_{\mathrm{E}} &= C_1\frac{\mathrm{APAR}}{\mathrm{daylength}}, \\
J_{\mathrm{C}} &= C_2 V_{\max}, \\
A_{\mathrm{gd}} &= \frac{J_{\mathrm{E}}+J_{\mathrm{C}}-
\sqrt{(J_{\mathrm{E}}+J_{\mathrm{C}})^2-4\theta J_{\mathrm{E}}J_{\mathrm{C}}}}
{2\theta\,\mathrm{daylength}},
\label{eq:Agd}
\end{align}
where $C_1$ and $C_2$ include temperature and conductance effects, APAR is absorbed photosynthetically active radiation, and $\theta$ controls co-limitation.

NeuralCrop provides neural representations for two quantities in this formulation. The optimal ratio of intercellular to atmospheric CO\textsubscript{2} concentration, $\lambda$, is normally found by solving
\begin{align}
A_{\mathrm{nd}} + (1-\mathrm{daylength}/24)R_{\mathrm{leaf}}
-p_a(g_c-g_{\min})(1-\lambda)/1.6 = 0,
\label{eq:lambda}
\end{align}
where $A_{\mathrm{nd}}$ is daily net photosynthesis, $R_{\mathrm{leaf}}$ is leaf respiration, $p_a$ is the atmospheric partial pressure of CO\textsubscript{2}, and $g_c$ and $g_{\min}$ are canopy conductances. An MLP emulates this iterative solution from day length, air temperature, and soil moisture.

A second MLP represents the maximum Rubisco capacity, $V_{\max}$, which is otherwise calculated as
\begin{align}
V_{\max} = \frac{N_{\mathrm{leaf}}-NC_{\mathrm{leaf,low}}C_{\mathrm{leaf}}}
{p\exp[k_{\mathrm{temp}}(T-25)]\,\mathrm{LAI}}\,\mathrm{daylength},
\label{eq:Vmax}
\end{align}
where $N_{\mathrm{leaf}}$ and $C_{\mathrm{leaf}}$ are leaf nitrogen and carbon, $NC_{\mathrm{leaf,low}}$ is the lower leaf nitrogen to carbon ratio, and $p$ and $k_{\mathrm{temp}}$ are empirical coefficients. The MLP uses day length, APAR, leaf nitrogen, and air temperature. Both networks are embedded in the differentiable photosynthesis calculation so that their parameters can be optimized through the simulated carbon dynamics.

\subsection{Crop carbon and water dynamics} \label{sec:crop_carbon_allocation}

Crop carbon allocation is governed by phenological development, organ priority, and water and nitrogen stress. Net primary productivity accumulates as biomass and is allocated among roots, leaves, storage organs, and a mobile carbon pool \cite{schaphoff2018lpjml4_s}:
\begin{align}
C_{\mathrm{root}} &= f_{\mathrm{root}}\,\mathrm{biomass}, \\
C_{\mathrm{leaf}} &= \frac{\mathrm{LAI}}{\mathrm{SLA}}, \\
C_{\mathrm{storage}} &= \mathrm{HI}(1-f_{\mathrm{root}})\,\mathrm{biomass}, \\
C_{\mathrm{pool}} &= \mathrm{biomass}-C_{\mathrm{root}}-C_{\mathrm{leaf}}-C_{\mathrm{storage}},
\label{eq:crop_allocation}
\end{align}
where $f_{\mathrm{root}}$ is the root allocation fraction, SLA is specific leaf area, and HI is the harvest index. NeuralCrop also provides a neural ordinary differential equation (NODE) representation based on the matrix formulation of Luo et al.~\cite{luo2022matrix_s}:
\begin{align}
\frac{d\mathbf{C}(t)}{dt}
=\mathbf{A}(\mathbf{C},t)\mu(t)
+\mathbf{B}(\mathbf{C},t)\boldsymbol{\xi}(t)\mathbf{K}(\mathbf{C},t)\mathbf{C}(t),
\label{eq:matrix}
\end{align}
where $\mathbf{C}(t)$ contains the crop carbon pools, $\mu(t)$ is the carbon input, and $\mathbf{A}$, $\mathbf{B}$, $\boldsymbol{\xi}$, and $\mathbf{K}$ describe allocation, transfer, environmental modification, and turnover. An MLP parameterizes the carbon pool tendencies from the current pools, net primary productivity, leaf area index, leaf nitrogen, and soil moisture \cite{chen2018neural_s}.

Crop respiration is the sum of root, storage, and mobile pool maintenance respiration and growth respiration:
\begin{align}
R_{\mathrm{crop}}
=R_{\mathrm{root}}+R_{\mathrm{storage}}+R_{\mathrm{pool}}+R_{\mathrm{growth}}.
\label{eq:crop_respiration}
\end{align}
NeuralCrop applies a bounded multiplicative correction,
\begin{align}
\widehat{R}_{\mathrm{crop}} &= m_R R_{\mathrm{crop}}, &
m_R &= 2\sigma\!\left(f_{\theta_R}(\mathbf{z}_R)\right),
\label{eq:neural_crop_respiration}
\end{align}
where $\sigma$ is the logistic function and $m_R$ lies between 0 and 2. The input vector $\mathbf{z}_R$ contains gross assimilation and leaf respiration, air and topsoil temperatures, plus carbon stored in roots and in the storage and mobile pools. Ecosystem respiration combines the corrected crop contribution with the process calculations of leaf and heterotrophic respiration.

Crop transpiration is determined by the balance between root water supply, $S_T$, and atmospheric demand, $D_T$:
\begin{align}
S_T &= e_{\max}w_r\left(1-\exp[-0.04C_{\mathrm{root}}]\right), \\
D_T &= \frac{(1-w_{\mathrm{can}})E_{\mathrm{eq}}\alpha_m}
{1+g_m\alpha_m/g_c},
\label{eq:transpiration_supply_demand}
\end{align}
where $w_r$ is root weighted soil water, $w_{\mathrm{can}}$ is the wet canopy fraction, $E_{\mathrm{eq}}$ is equilibrium evapotranspiration, and $g_c$ is canopy conductance. A second bounded correction modifies the water supply used by the transpiration calculation:
\begin{align}
\widehat{S}_T &= \min\!\left[m_T\min(S_T,D_T),D_T\right], &
m_T &= 2\sigma\!\left(f_{\theta_T}(\mathbf{z}_T)\right).
\label{eq:neural_crop_transpiration}
\end{align}
The input vector $\mathbf{z}_T$ contains day length, equilibrium evapotranspiration, absorbed radiation fraction, canopy wetness, canopy conductance, root carbon, and root weighted soil moisture. The corrected flux remains constrained by atmospheric demand and the plant available water in each soil layer before it enters the soil water balance and total evapotranspiration.

\subsection{Soil carbon cycles}

The litter cycle contains aboveground leaf and stem pools and a belowground root pool. Soil organic matter is represented by fast and slow pools in each of five soil layers~\cite{sitch2003evaluation}. NeuralCrop expresses the litter and soil carbon balances as ordinary differential equations:
\begin{align}
\frac{dC_{\mathrm{litter},l}}{dt}
&=A_{\mathrm{litter},l}C_{\mathrm{residue}}
-\left[1-\exp(-k_{\mathrm{litter},10}f_{\mathrm{litter},l})\right]C_{\mathrm{litter},l},
&&l=1,2,3,
\label{eq:litter}\\
\frac{dC_{\mathrm{fast},l}}{dt}
&=A_{\mathrm{fast},l}\mu_C
-\left[1-\exp(-k_{\mathrm{fast},10}f_{\mathrm{fast},l})\right]C_{\mathrm{fast},l},
&&l=1,\ldots,5,
\label{eq:fast}\\
\frac{dC_{\mathrm{slow},l}}{dt}
&=A_{\mathrm{slow},l}\mu_C
-\left[1-\exp(-k_{\mathrm{slow},10}f_{\mathrm{slow},l})\right]C_{\mathrm{slow},l},
&&l=1,\ldots,5,
\label{eq:slow}
\end{align}
where $C_{\mathrm{residue}}$ is residue carbon, $\mu_C$ is decomposed litter carbon entering the soil, $A$ denotes partitioning among pools and layers, $k_{10}$ is the reference decomposition rate at $10\,^{\circ}\mathrm{C}$, and $f$ is the environmental response. Aboveground litter decomposition depends on air temperature and soil moisture, whereas belowground litter and soil organic matter depend on soil temperature and moisture.

The process formulation represents the environmental response with
\begin{align}
f(T,\vartheta)
=\exp\!\left[308.56\left(\frac{1}{56.02}-\frac{1}{T+46.02}\right)\right]
\left(0.0402-5.005\vartheta^3+4.269\vartheta^2+0.719\vartheta\right),
\label{eq:response}
\end{align}
where $T$ is air or soil temperature and $\vartheta$ is soil moisture. NeuralCrop can replace the response functions of the litter, fast, and slow pools with MLPs that take temperature and soil moisture as inputs.

\subsection{Soil water cycle} \label{sec:soil_water_cycle} 

Soil water is represented by five hydrologically active layers and updated with a daily bucket model \cite{rost2008agricultural, schaphoff2018lpjml4_s}:
\begin{align}
W_t=W_{t-1}+\Delta t\left[(P+M+I_{\mathrm{rr}}-E_I)-R-(E_S+E_T)-p\right],
\label{eq:soilwater}
\end{align}
where $W$ is the water stored in a soil layer, $P$ is precipitation, $M$ is snowmelt, $I_{\mathrm{rr}}$ is irrigation, $E_I$ is canopy interception loss, $R$ is runoff, $E_S$ is soil evaporation, $E_T$ is crop transpiration, and $p$ is percolation. With a daily time step, this balance can be written as
\begin{align}
\frac{dW}{dt}=(P+M+I_{\mathrm{rr}}-E_I)-R-(E_S+E_T)-p.
\label{eq:soilwater_ode}
\end{align}

\begin{figure}
\centering
\includegraphics[width=1\textwidth]{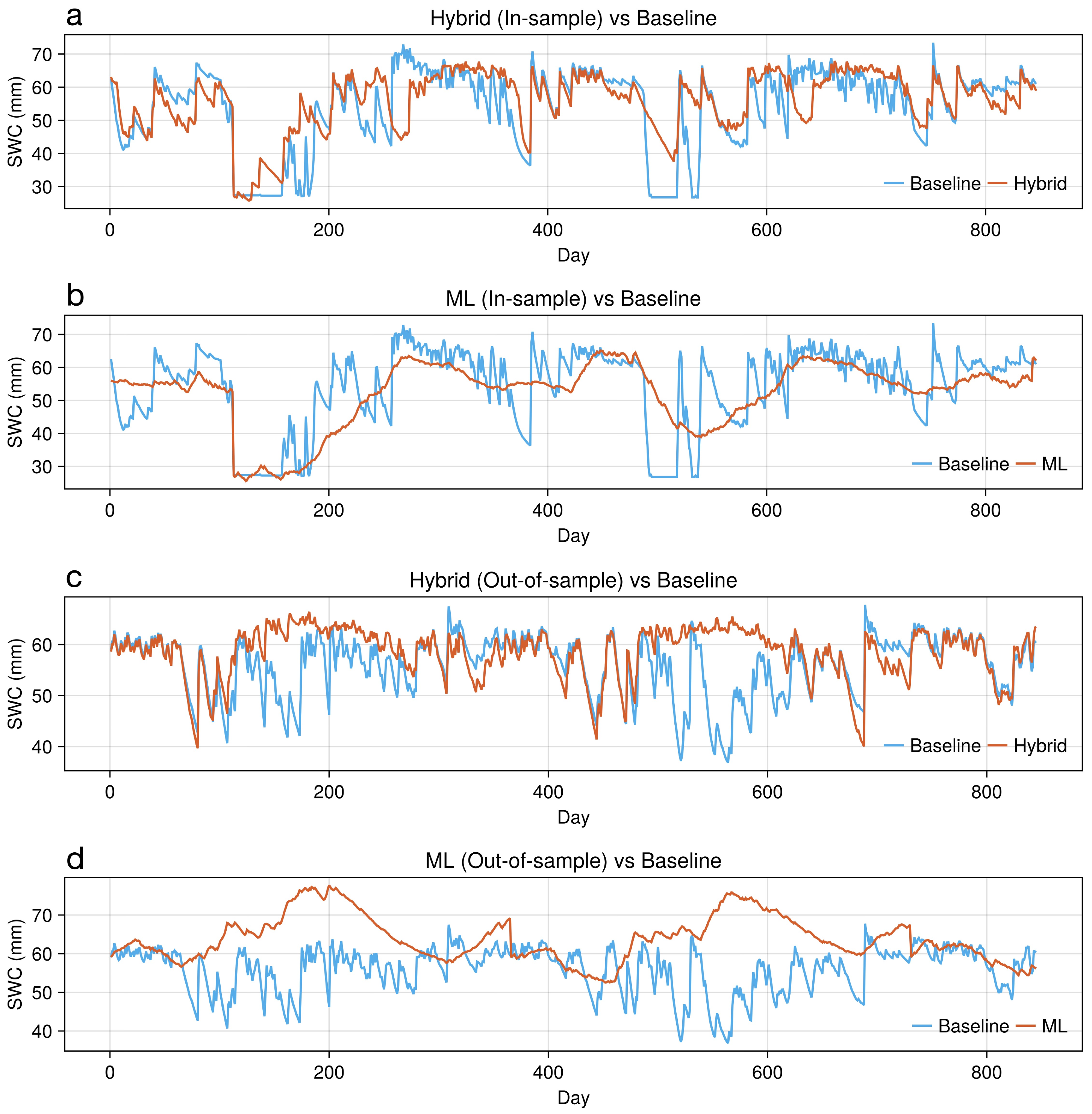}
\caption{Comparison of soil water content (SWC) simulated by hybrid and pure machine learning models under in-sample and out-of-sample scenarios, with LPJmL-simulated SWC used as baseline. Panels a and b show in-sample performance for the hybrid and pure machine learning models, respectively, while panels c and d present their out-of-sample performance.}\label{fig_swc_hybrid_ML}
\end{figure}

NeuralCrop implements this balance in a differentiable form and provides two neural representations. In this hybrid representation, MLPs replace the snowmelt and soil evaporation terms while the remaining fluxes retain their process formulations. In the fully learned representation, an MLP represents the complete daily soil water tendency on the right hand side of Eq.~\ref{eq:soilwater_ode}. The comparison in Fig.~\ref{fig_swc_hybrid_ML} shows that retaining the process constraints better preserves rapid soil water fluctuations, whereas the fully learned representation produces smoother trajectories and larger deviations under conditions outside the training sample.

\clearpage

\section{Data} \label{sec:data}

\subsection{Initial conditions}

Before simulating managed croplands, process-based vegetation models usually need spin-up simulations to develop equilibrium vegetation and soil states. LPJmL involves a two-stage spin-up process. The first spin-up simulation typically runs for 5000 years until all system states (i.e., soil carbon, nitrogen, and water pools) reach equilibrium. The second spin-up simulation typically spans 390 years to introduce the impacts of land-use change on these pools driven by historical human activities \cite{minoli2022global_s}. Since NeuralCrop does not simulate natural vegetation, it cannot perform this spin-up procedure prior to crop simulations. We therefore perform the spin-up in LPJmL and initialize the soil carbon, nitrogen, and water pools in NeuralCrop from LPJmL simulations to ensure physically consistent initial conditions.

\subsection{Pre-training data}

Due to the limited availability of large-scale high-resolution observations for cropland carbon, water, and nitrogen dynamics, NeuralCrop is pre-trained on its fully process-based simulations (Table \ref{pre_training_variables}). This pre-training step allows NeuralCrop to learn prior process knowledge and provides a strong initialization for subsequent training under data-scarce conditions. The model inputs are shown in Table \ref{lpjml_input}. 

\begin{table}[ht]
\centering
\caption{Training data generated by the fully process-based simulations.}
\label{pre_training_variables} 
\begin{tabular}{lcccc}
\toprule
\textbf{Variable} & \textbf{Symbol}  & \textbf{Dimension} & \textbf{Units}  & \textbf{Time} \\
\midrule
Inter-cellular to atmospheric CO\textsubscript{2} ratio  & $\lambda$  & 1 & -    & daily   \\
Maximum Rubisco capacity       & $V_{\text{max}}$   & 1 & $\mathrm{g\,C\,m^{-2}}$    & daily   \\
Gross primary productivity       & GPP   & 1 & $\mathrm{g\,C\,m^{-2}}$   & daily   \\
Crop carbon pools      & $C_{\text{i}}$   & 4  & $\mathrm{g\,C\,m^{-2}}$    & daily   \\
Crop respiration       & $R_{\mathrm{crop}}$   & 1 & $\mathrm{g\,C\,m^{-2}}$   & daily   \\
Crop transpiration       & $T_{\mathrm{crop}}$   & 1 & $\mathrm{mm}$   & daily   \\
Litter carbon pools      & $C_{\text{litter(l)}}$   & 3   & $\mathrm{g\,C\,m^{-2}}$    & daily   \\
Soil fast carbon pools      & $C_{\text{fast(l)}}$   & 5   & $\mathrm{g\,C\,m^{-2}}$    & daily   \\
Soil slow carbon pools      & $C_{\text{slow(l)}}$   & 5   & $\mathrm{g\,C\,m^{-2}}$    & daily   \\
Soil water pools      & $W_{\text{(l)}}$   & 5  & $\mathrm{mm}$  & daily   \\

\bottomrule
\end{tabular}
\end{table}

\begin{table}
\centering
\caption{Model inputs.}
\label{lpjml_input} 
\begin{tabular}{lccc}
\toprule
\textbf{Type} & \textbf{Variable} & \textbf{Units} & \textbf{Time} \\
\midrule
Climate  & Temperature  & $^{\circ}\mathrm{C}$  & daily   \\
Climate  & Precipitation   & $\mathrm{mm}$    & daily   \\
Climate  & Net longwave radiation    & $\mathrm{W/m^2}$  & daily   \\
Climate  & Downward shortwave radiation   & $\mathrm{W/m^2}$   & daily   \\
Climate  & CO\textsubscript{2}   & $\mathrm{ppmv}$ & annual   \\
Soil  & Soil texture   & -   & static   \\
Soil  & Soil pH   & -  & static   \\
Management  & Land use   & -   & annual   \\
Management  & Nitrogen fertilizer   & $\mathrm{g\,N\,m^{-2}}$   & annual   \\
Management  & Tillage   & -   & annual   \\
Management  & Residue on field   & -    & annual   \\
Management  & Sowing date   & -    & static   \\
Management  & Heat unit requirements   & -   & static   \\

\bottomrule
\end{tabular}
\end{table}

\subsection{Fine-tuning data} \label{SI_subsection_fine-tuinign_data}

During the fine-tuning stage, the training of NeuralCrop is focused on optimizing the ML components with real-world observations. The fine-tuning data were obtained from the global flux network FLUXNET, the American flux network AmeriFlux, and the European Integrated Carbon Observation System (ICOS), which provide site-level daily crop and soil eddy covariance flux as well as meteorological forcing measurements. Table \ref{fluxnet_data} shows the observational data used during the fine-tuning stage. Only the neural components linked to the available flux observations are fine-tuned, while the remaining components are kept fixed.

\begin{table}[h]
\centering
\caption{Flux network observational data.}
\label{fluxnet_data} 
\begin{tabular}{lcccc}
\toprule
\textbf{Type} & \textbf{Variable} & \textbf{Symbol} & \textbf{Units} & \textbf{Time} \\
\midrule
Feature  & Air temperature   & $\mathrm{TA}_\mathrm{F}$    & $^{\circ}\mathrm{C}$  & daily   \\
Feature  & Precipitation   & $\mathrm{P}_\mathrm{F}$    & $\mathrm{mm}$    & daily   \\
Feature  & Downward shortwave radiation  & $\mathrm{SW}_{\mathrm{IN\_F}}$    & $\mathrm{W/m^2}$  & daily   \\
Feature  & Downward longwave radiation   & $\mathrm{LW}_{\mathrm{IN\_F}}$    & $\mathrm{W/m^2}$   & daily   \\
Feature  & Upward longwave radiation   & $\mathrm{LW}_{\mathrm{OUT}}$    & $\mathrm{W/m^2}$   & daily   \\
Feature  & CO\textsubscript{2}   & $\mathrm{CO2}_{\mathrm{F\_MDS}}$   & $\mu\mathrm{mol}\,\mathrm{mol^{-1}}$   & daily   \\
Label  & Gross primary productivity   & $\mathrm{GPP}_{\mathrm{NT\_VUT\_REF}}$    & $\mathrm{g\,C\,m^{-2}}$   & daily   \\
Label  & Ecosystem Respiration   & $\mathrm{RECO}_{\mathrm{NT\_VUT\_REF}}$    & $\mathrm{g\,C\,m^{-2}}$   & daily   \\
Label  & Evapotranspiration   & derived from $\mathrm{LE}_{\mathrm{F\_MDS}}$    & $\mathrm{mm\,d^{-1}}$   & daily   \\

\bottomrule
\multicolumn{5}{l}{\footnotesize{ET is derived from latent heat flux using temperature-dependent latent heat of vaporization.}} \\
\end{tabular}
\end{table}

\subsection{Data pre-processing} \label{sec:data_pre-processing}

The flux networks provide both daytime and nighttime partitioned GPP and RECO~\cite{reichstein2005separation_s, pastorello2020fluxnet2015_s}. The nighttime method estimates RECO by fitting a respiration-temperature function using nighttime data and derives GPP as the difference between modeled RECO and daytime net ecosystem exchange (NEE). Because photosynthesis is negligible at night, RECO is approximately equal to NEE. The daytime method fits a light-response function to estimate GPP and a respiration-temperature function for RECO using daytime and nighttime data. As a result, both GPP and RECO are inferred from NEE, which introduces coupled uncertainties. Therefore, we use nighttime partitioned GPP and RECO for fine-tuning. The nighttime partitioned GPP and RECO may contain negative values. We replace negative GPP and RECO with their daytime counterparts to avoid introducing an additional bias during fine-tuning. 

Evapotranspiration is derived from gap-filled latent heat flux $\mathrm{LE}_{\mathrm{F\_MDS}}$ as $\mathrm{ET}=\mathrm{LE}\times86400/(2.501\times10^6-2361T)$, where $T$ is daily air temperature in $^\circ$C and the resulting unit is mm~d$^{-1}$. Finite negative ET values are retained to preserve the source product. Each physical site receives equal total weight in training, and seasons within a site receive equal weight.

\clearpage

\section{Training} \label{sec:training}

\subsection{Input data normalization}

All input features are standardized before being joined together and passed to the neural networks. We standardize climate forcings to zero mean and unit variance to improve training stability. Specifically, for each climate forcing variable \( x \), we compute the standardized value \( \tilde{x} \) as:
\begin{align}
\tilde{x} = \frac{x - \mu}{\sigma},
\end{align}
where \( \mu \) is the mean and \( \sigma \) is the standard deviation computed from the 20-year historical climatology. For input features generated by NeuralCrop itself, we rescale them to the \([0, 1]\) range using their physical upper and lower bounds. Here, for each state variable \( x \), the normalized value \( \tilde{x} \) is computed as:
\begin{align}
\tilde{x} = \frac{x - x_{\text{lower}}}{x_{\text{upper}} - x_{\text{lower}}},
\end{align}
where \( x_{\text{lower}} \) and \( x_{\text{upper}} \) denote the physical lower and upper limits of the variable, respectively.

\subsection{Optimizer settings}

All NeuralCrop models were pre-trained using the AdamW optimizer \cite{loshchilov2017decoupled_s}. We used $\beta_1 = 0.9$, $\beta_2 = 0.999$, and $\epsilon = 10^{-6}$ based on empirical evidence. The initial pre-training learning rate was set to $1 \times 10^{-3}$. A sinusoidal exponential decay scheme ($\text{SinExp}$) was employed to dynamically adjust the learning rate throughout training, oscillating between $1 \times 10^{-3}$ and $1 \times 10^{-5}$ with a decay factor of $0.975$ and a 20-epoch cycle. The only exception to these hyperparameters in fine-tuning was that the initial learning rate was set to $1 \times 10^{-4}$. The grouped site-holdout models were trained for 150 epochs with three random seeds.

\subsection{Loss function} \label{sec:loss_function}

All NeuralCrop models are trained in two stages (pre-training and fine-tuning) with different loss functions, where all loss terms are defined in the general form of mean squared error (MSE): 
\begin{align}
\mathrm{MSE} = \frac{1}{T} \sum_{t=1}^{T} \left( \hat{y}_t - y_t \right)^2, \label{eq:mse}
\end{align}
where $\hat{y}_t$ is the model simulation at time $t$, $y_t$ is the corresponding observation, 
and $T$ is the total number of time steps.

During the pre-training stage, we employed a combination of two loss types (i.e., accuracy and mass balance) to minimize two kinds of discrepancies between the hybrid and fully process-based outputs. The loss term of accuracy accounts for the distance between predictions and reference data, and the loss term of mass balance constrains the carbon conservation during crop carbon allocation. The pre-training loss function has two kinds of loss types:
\begin{align}
\mathcal{L}_{\text{Pre-training}} = \sum_{i \in \mathcal{T}} \beta_{\text{accuracy}, i} \mathcal{M}_{\text{Accuracy}, i} + \beta_{\text{mass}} \mathcal{M}_{\text{MassBalance}}, \label{eq:loss1}
\end{align}
where $\mathcal{T}$ denotes the set of loss terms of accuracy (i.e., $\lambda$, $V_{\text{max}}$, GPP, carbon pools, soil water pools), $\mathcal{M}_{\text{Accuracy}, i}$ is the MSE between NeuralCrop prediction and reference data of variable $i$ with corresponding loss scales $\beta_{\text{accuracy}, i}$, $\mathcal{M}_{\text{MassBalance}}$ is the MSE between the sum of crop carbon and accumulated total crop biomass with a loss scale $\beta_{\text{mass}}$.

After pre-training, NeuralCrop was fine-tuned with eddy-covariance observations. Only components linked to the available direct targets were updated (Section~\ref{SI_subsection_fine-tuinign_data}). The fine-tuning loss is
\begin{align}
\mathcal{L}_{\text{Fine-tuning}} = \beta_{\text{gpp}} \mathcal{M}_{\text{GPP}} + \beta_{\text{reco}} \mathcal{M}_{\text{RECO}} + \beta_{\text{et}} \mathcal{M}_{\text{ET}}, \label{eq:loss2}
\end{align}
where $\mathcal{M}_{\text{GPP}}$, $\mathcal{M}_{\text{RECO}}$, and $\mathcal{M}_{\text{ET}}$ denote MSE for GPP, RECO, and ET, scaled by $\beta_{\text{gpp}}$, $\beta_{\text{reco}}$, and $\beta_{\text{et}}$, respectively.

\subsection{Training rollouts} \label{sec:training_rollouts}

\begin{figure}[H]
\centering
\includegraphics[width=0.72\textwidth]{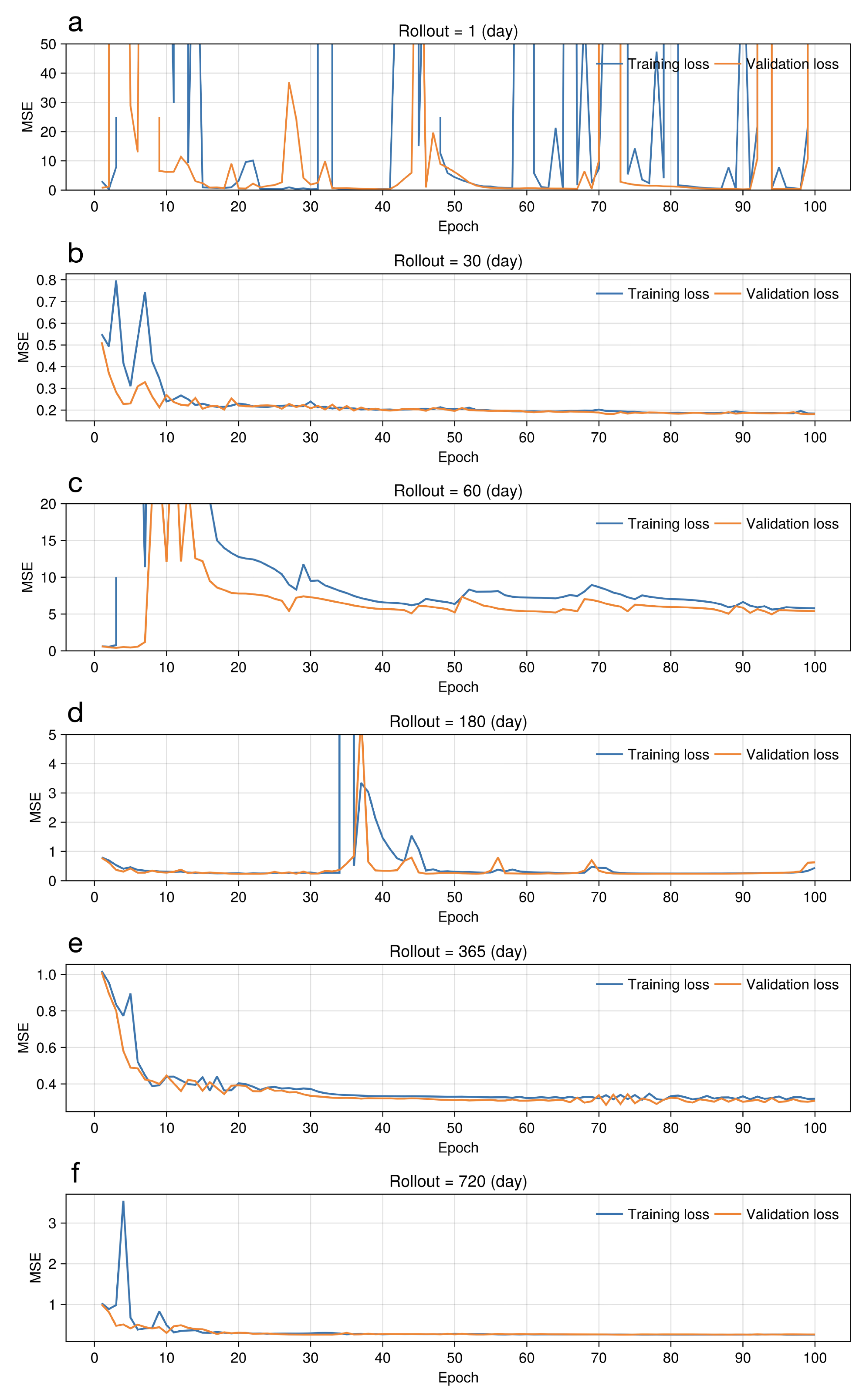}
\caption{NeuralCrop models with identical model parameters trained on rollouts of 1, 30, 60, 180, 365, and 720 days (a-f).}\label{fig_rollouts}
\end{figure}

To explore the benefits of training with different rollouts (i.e., how many simulation days to update ML parameters), we trained the NeuralCrop models with identical model parameters on rollouts of 1, 30, 60, 180, 365, and 720 days (Fig.~\ref{fig_rollouts}). We find that longer rollouts yield more stable loss trajectories. NeuralCrop trained on shorter rollouts, especially 1-day window, exhibits large oscillations in both training and validation loss during the entire training period. The 365- and 720-day rollouts show similarly smooth convergence. We select 365 days for pre-training to limit computational cost. Each pre-training rollout begins at sowing and spans the crop season. Site-level fine-tuning instead uses a season-specific window, with only valid flux-observation days contributing to the loss.

\subsection{Output rescaling}

All variables predicted by neural networks during training are rescaled using their historical minimum and maximum values over the training years, which prevents variables with large magnitudes from dominating the training. Specifically, for each predicted variable \( x \), the normalized value \( \tilde{x} \) is computed as:
\begin{align}
\tilde{x} = \frac{x - x_{\text{min}}}{x_{\text{max}} - x_{\text{min}}},
\end{align}
where \( x_{\text{min}} \) and \( x_{\text{max}} \) denote the minimum and maximum values of the variable, respectively.

\subsection{Training resources} \label{sec:training_resources}

All training and inference of NeuralCrop were conducted using a single NVIDIA H100 GPU with 80GB of memory.

\begin{figure}[H]
\centering
\includegraphics[width=0.75\textwidth]{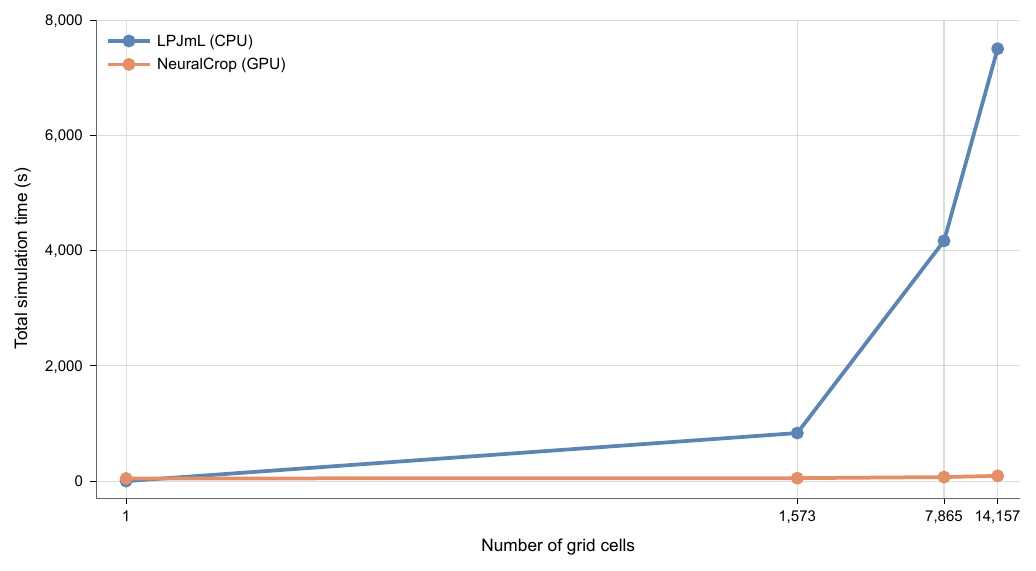}
\caption{Inference time of GPU-accelerated NeuralCrop and CPU-based LPJmL simulating for a $0.5^\circ \times 0.5^\circ$ spatial resolution with daily time steps over a 20-year simulation period (7300 days) for different numbers of grid cells. NeuralCrop was running on one single NVIDIA H100 GPU with 80GB of memory, and LPJmL was running on a CPU node equipped with two AMD EPYC 9554 processors (128 cores in total) and 768 GB of memory.}\label{fig_time_efficiency}
\end{figure}

We compared the efficiency of inference time between GPU-accelerated NeuralCrop and CPU-based LPJmL by running them on a $0.5^\circ \times 0.5^\circ$ spatial resolution with daily time steps over a 20-year simulation period (7300 days) for different numbers of grid cells (Table~\ref{table_time_efficiency} and Figure~\ref{fig_time_efficiency}). The LPJmL was running on a CPU node equipped with two AMD EPYC 9554 processors (128 cores in total) and 768 GB of memory. The results show that LPJmL takes over 7,500 seconds (about 2 hours) to simulate 14,157 grid cells, whereas NeuralCrop completes the corresponding simulation in 91 seconds, over 82 times faster. Similar gains appear at smaller grid cells, with 60 times faster for 7,865 grid cells and 16 times faster for 1,573 grid cells. Although GPU overhead makes the one-cell simulation slower, the efficiency increases rapidly with scale, reflecting the powerful parallel capacity of the GPU. This allows NeuralCrop to achieve competitive speed compared to traditional GGCM running on multiple CPU cores when doing global and daily simulations.

\begin{table}[h]
\centering
\caption{Inference time of GPU-accelerated NeuralCrop and CPU-based LPJmL simulating for a $0.5^\circ \times 0.5^\circ$ spatial resolution with daily time steps over a 20-year simulation period (7300 days) for different numbers of grid cells. NeuralCrop was running on one single NVIDIA H100 GPU with 80GB of memory, and LPJmL was running on a CPU node equipped with two AMD EPYC 9554 processors (128 cores in total) and 768 GB of memory.}
\label{table_time_efficiency}
\begin{tabular}{lccccc}
\toprule
\textbf{Grid cells} & \textbf{LPJmL} & \textbf{NeuralCrop} & \textbf{Times (speed up)}\\
\midrule
1   & 0.50 s  & 46.01 s & 0.011x  \\
1573   & 833.99 s  & 51.24 s & 16.27x  \\
7865   & 4169.94 s  & 70.02 s & 59.55x  \\
14157   & 7505.89 s  & 91.16 s & 82.33x  \\
\bottomrule
\end{tabular}
\end{table}

\clearpage

\section{Evaluation metrics}\label{sec:evaluation_metrics}

We use complementary metrics to measure optimization or error magnitude. Throughout, $y_t$ denotes an observation, $\hat y_t$ a simulation, and $T$ the number of valid time steps.

\subsection{Pearson correlation coefficient} \label{sec:Pearson_correlation_coefficient}

Pearson correlation coefficient ($r$) calculates the linear relationship between model simulations and observations. It is defined as:
\begin{align}
r = \frac{\sum_{t=1}^{T} (y_t - \bar{y})(\hat{y}_t - \bar{\hat{y}})}{\sqrt{\sum_{t=1}^{T} (y_t - \bar{y})^2} \sqrt{\sum_{t=1}^{T} (\hat{y}_t - \bar{\hat{y}})^2}}, \label{eq:r}
\end{align}
where $\hat{y}_t$ and $y_t$ represent the simulated and observed values at time $t$, respectively, while $\bar{\hat{y}}$ and $\bar{y}$ denote their mean values. It ranges from $-1$ to $+1$, with higher values indicating stronger agreement in the temporal variability. In particular, $r$ is used to assess the ability of models to reproduce seasonal and interannual variations.

\subsection{Root mean square error}\label{sec:rmse}

Root mean square error (RMSE) is a widely used metric to evaluate the accuracy of model predictions against observations. It is defined as the square root of the average of the squared differences between predicted values and observed values:
\begin{align}
RMSE = \sqrt{\frac{1}{T} \sum_{t=1}^{T} (\hat{y}_t - y_t)^2}, \label{eq:rmse}
\end{align}
where $\hat{y}_t$ is the model simulation at time $t$, $y_t$ is the corresponding observation, 
and $T$ is the total number of time steps. A lower RMSE value indicates better agreement between model simulations and observations.

For daily site fluxes, RMSE is calculated without detrending. Within each held-out season, we calculate
\begin{align}
\mathrm{nRMSE}_{\mathrm{site}}
=\frac{\mathrm{RMSE}}{\sigma_{y}},
\end{align}
where $\sigma_y$ is the sample standard deviation of the observed daily values.

For regional yields, simulated and reported annual series are linearly detrended separately. Within each administrative unit, RMSE is calculated between the detrended series and normalized as
\begin{align}
\mathrm{nRMSE}_{\mathrm{yield}}
=\frac{\mathrm{RMSE}}{\sigma_{y'}},
\end{align}
where $\sigma_{y'}$ is the standard deviation of the detrended reported yields.

\subsection{Mean absolute error}

Mean absolute error is
\begin{align}
\mathrm{MAE}=\frac{1}{T}\sum_{t=1}^{T}|\hat y_t-y_t|.
\end{align}
Here, $\hat y_t$ and $y_t$ are the simulated and observed values, respectively, and $T$ is the number of paired observations. Lower MAE indicates closer agreement between simulations and observations.

\clearpage

\section{Independent site evaluation} \label{sec:site-level_validation}
\subsection{Sites and site groups}

\begin{figure}[H]
\centering
\includegraphics[width=1.0\textwidth]{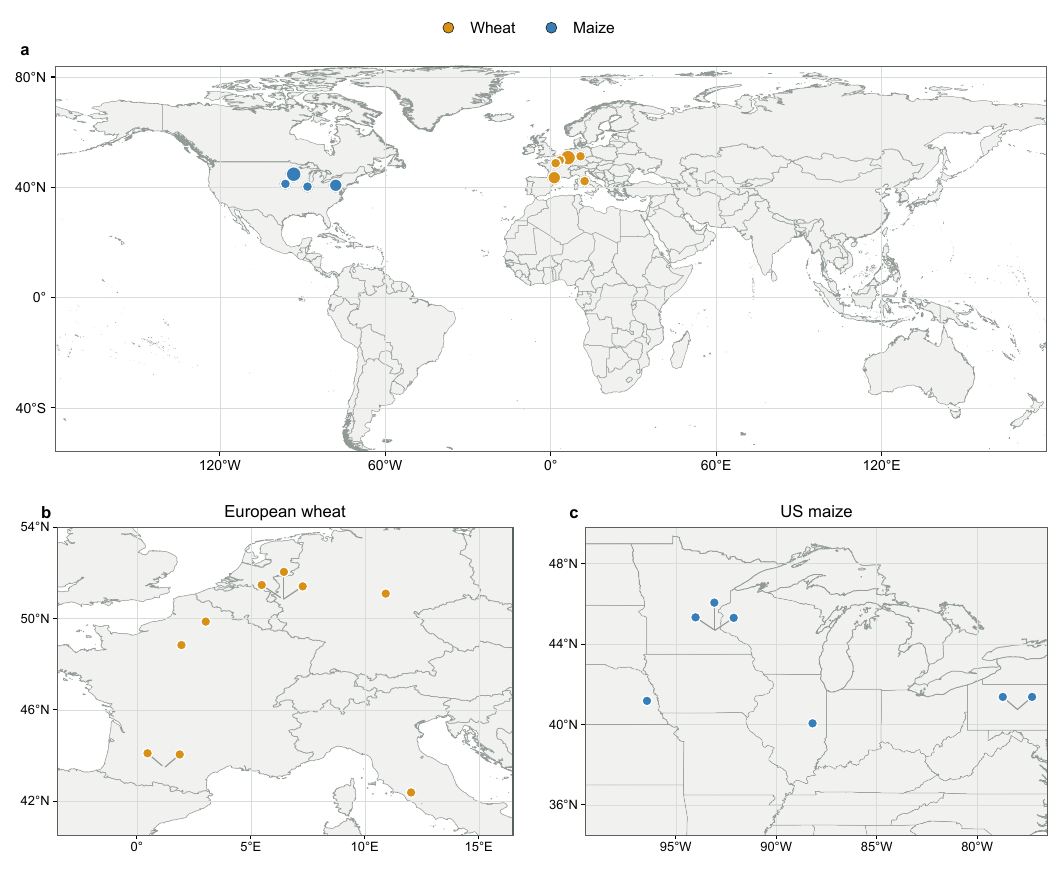}
\caption{Distribution of flux tower sites used in the grouped site held-out experiment. \textbf{a}, Global locations of the independent site groups. Orange and blue markers denote European wheat and US maize groups, respectively. Marker size indicates the number of physical sites in each group. \textbf{b,c}, Enlarged the physical wheat and maize tower sites. Markers for nearby sites in the same group are displaced for visibility. Site names and season counts are given in Table~\ref{table:station_groups}. The evaluation contains 10 independent site groups, 16 physical sites, and 50 crop seasons.}\label{fig:station_distribution}
\end{figure}

The site evaluation covers wheat and maize. After quality control, the dataset comprises 50 crop seasons from 16 flux tower sites. Sites located near one another are assigned to one site group, ensuring that closely related locations do not appear in both training and evaluation. This procedure defines six independent wheat site groups (29 seasons from nine sites) and four maize site groups (21 seasons from seven sites; Table~\ref{table:station_groups}). Their spatial distribution is shown in Fig.~\ref{fig:station_distribution}.

\begin{table}[h]
\centering
\caption{Flux tower sites and independent site groups used for the holdout evaluation. Season counts are the crop seasons remaining after quality control.}
\label{table:station_groups}
\begin{tabular}{llllr}
\toprule
\textbf{Crop} & \textbf{Group} & \textbf{Sites} & \textbf{Grouping basis} & \textbf{Seasons}\\
\midrule
Wheat & W01 & DE-Geb & standalone site & 4\\
Wheat & W02 & DE-RuM, DE-RuS, DE-Seh & nearby sites & 11\\
Wheat & W03 & FR-Aur, FR-Lam & nearby sites & 8\\
Wheat & W04 & FR-EM2 & standalone site & 2\\
Wheat & W05 & FR-Gri & standalone site & 2\\
Wheat & W06 & IT-CA2 & standalone site & 2\\
Maize & M01 & US-Ne3 & standalone site & 2\\
Maize & M02 & US-Ro1, US-Ro3, US-Ro5 & nearby sites & 7\\
Maize & M03 & US-UC1, US-UC2 & nearby sites & 9\\
Maize & M04 & US-UiG & standalone site & 3\\
\bottomrule
\end{tabular}
\end{table}

\subsection{Site group holdout design}

For each crop, one site group is withheld at a time. All seasons from every site in that group are excluded from fine-tuning. The remaining groups provide the training data. Each fold is fine-tuned for 150 epochs with three random seeds. The evaluated targets are daily GPP, RECO, and ET. Default and hybrid simulations use identical forcings, initial states, and crop calendars. The default simulation is the process-only configuration. The hybrid simulation uses the same process framework but activates the three trained neural components for GPP, crop respiration, and crop transpiration. For each ablation, one of these components is disabled during inference while all other components remain unchanged. The GPP component comprises the separate $\lambda$ and $V_{\max}$ MLPs, which are disabled together in the GPP-off configuration. The ablations therefore measure sensitivity to individual components rather than the performance of separately retrained models.

The 10 folds and three random seeds yield 30 fine-tuned models. Metrics are calculated only for seasons in the site group held out from training. For each bootstrap resample, we draw six wheat groups or four maize groups with replacement. A group selected multiple times contributes its group-level value once for each draw. The 2.5th and 97.5th percentiles of the resampled mean changes define the 95\% intervals.

\subsection{Results}

\begin{figure}[H]
\centering
\includegraphics[width=1.0\textwidth]{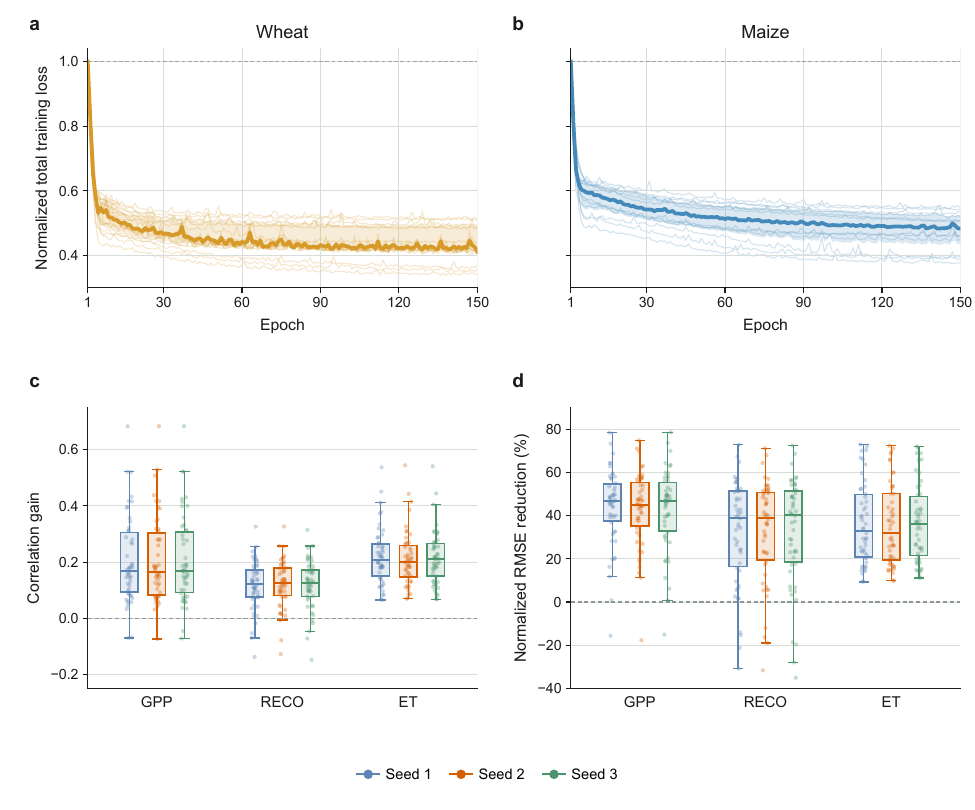}
\caption{Training loss trajectories and sensitivity to random seed for the grouped site holdout evaluation. \textbf{a,b}, Normalized training objective over 150 epochs for wheat and maize. Thin curves show individual fold and seed runs, solid curves show the median at each epoch, and shaded bands show the interquartile range. \textbf{c,d}, Differences in correlation and nRMSE between the default and hybrid simulations for GPP, RECO, and ET in seasons excluded from training. Colors denote the three random seeds, and points represent individual seasons.}
\label{fig:training_seed_sensitivity}
\end{figure}

The training loss trajectories across all three seeds are reported in Fig.~\ref{fig:training_seed_sensitivity}. Results with equal weight assigned to each site group are shown in Table~\ref{table:site_holdout_results}. For wheat, mean correlation gains range from 0.114 to 0.194, and nRMSE reductions range from 32.1\% to 51.0\%. For maize, correlation gains range from 0.120 to 0.266, and nRMSE reductions range from 18.0\% to 55.9\%. All 95\% bootstrap intervals for the paired mean changes exclude zero, indicating consistent improvements across the evaluated site groups.

\begin{table}[h]
\centering
\small
\caption{Performance at sites excluded from training. The table reports mean performance metrics for the default and hybrid simulations, assigning equal weight to each site group, together with the paired mean improvement and its 95\% bootstrap interval. Positive improvement indicates higher correlation or lower nRMSE for the hybrid simulations than for the default simulations.}
\label{table:site_holdout_results}
\setlength{\tabcolsep}{4pt}
\begin{tabular}{llrrrr}
\toprule
Crop & Target/metric & Default & Hybrid & Improvement & 95\% CI\\
\midrule
Wheat & GPP $r$ & 0.716 & 0.896 & +0.180 & [0.102, 0.293]\\
Wheat & RECO $r$ & 0.763 & 0.877 & +0.114 & [0.084, 0.143]\\
Wheat & ET $r$ & 0.681 & 0.875 & +0.194 & [0.163, 0.226]\\
Wheat & GPP nRMSE & 0.995 & 0.488 & 0.507 & [0.353, 0.648]\\
Wheat & RECO nRMSE & 0.677 & 0.390 & 0.287 & [0.163, 0.410]\\
Wheat & ET nRMSE & 0.778 & 0.528 & 0.249 & [0.181, 0.314]\\
Maize & GPP $r$ & 0.691 & 0.901 & +0.210 & [0.131, 0.281]\\
Maize & RECO $r$ & 0.645 & 0.766 & +0.120 & [0.043, 0.197]\\
Maize & ET $r$ & 0.528 & 0.795 & +0.266 & [0.226, 0.307]\\
Maize & GPP nRMSE & 0.682 & 0.483 & 0.199 & [0.146, 0.240]\\
Maize & RECO nRMSE & 0.661 & 0.542 & 0.119 & [0.026, 0.185]\\
Maize & ET nRMSE & 0.756 & 0.333 & 0.422 & [0.280, 0.601]\\
\bottomrule
\end{tabular}
\end{table}

\clearpage

\clearpage
\section{Regional yield simulation} \label{sec:cropping_regions}

\subsection{European wheat}
\subsubsection{Subnational crop statistics} \label{sec:subnational_data}

We use the harmonized EU subnational crop statistics dataset \cite{ronchetti2024harmonized_sup}, compiled from National Statistical Institutes and EUROSTAT. It contains production, area, and yield records for soft, durum, and total wheat across multiple NUTS levels. The analysis uses wheat records from 530 administrative units over 1981--2019. Reported wheat yields are converted from the standard EU humidity of 14\% to dry matter before analysis:
\begin{align}
Y_{\mathrm{DM}} = Y_{\mathrm{HM}} \times (1 - h), \label{eq:standardized_yield}
\end{align}
where $Y_{\mathrm{DM}} $ is the yield at 0\% humidity, $Y_{\mathrm{HM}}$ is the yield at the standard humidity, and $h$ is the standard humidity fraction.

\subsubsection{Pre-training}

In the European wheat cropping regions, we pre-train NeuralCrop against outputs generated by its process-based simulations from 2016-1-1 to 2017-12-31 for winter wheat under rain-fed and irrigated management and outputs from 2016-1-1 to 2016-12-31 for spring wheat under both management systems. The training setup is described in Sections~\ref{sec:data} and~\ref{sec:training}.

\begin{figure}[H]
\centering
\includegraphics[width=1.0\textwidth]{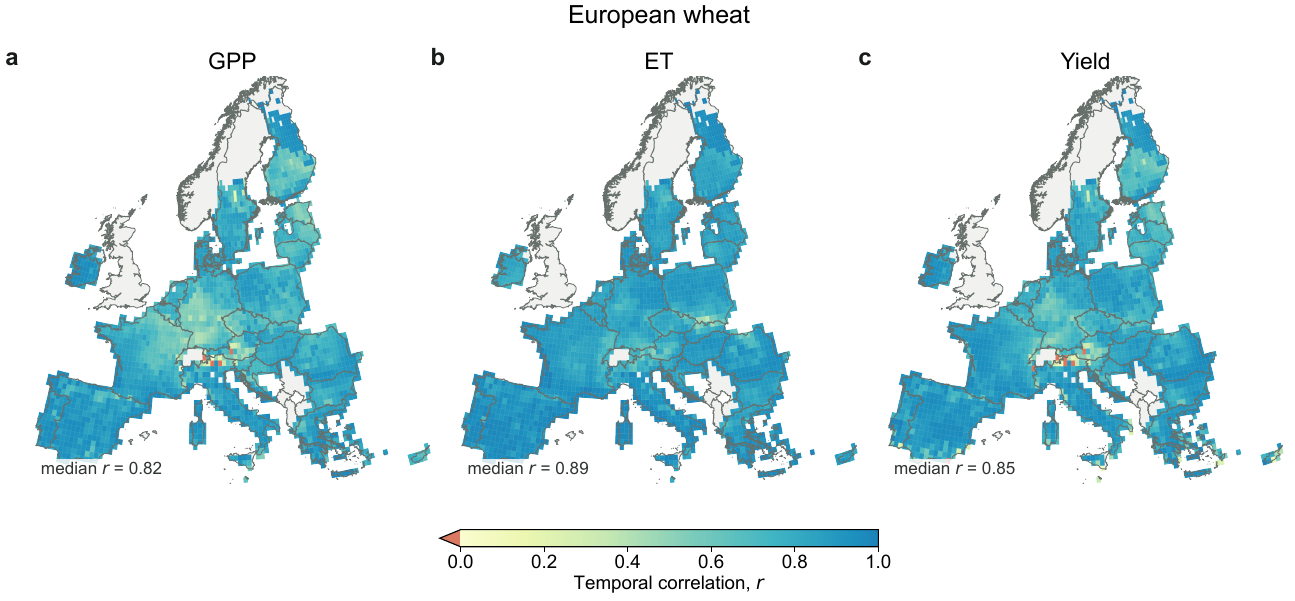}
\caption{Spatial distribution of grid-cell temporal correlations between the default and pre-trained simulations for European wheat regions at 0.5$^\circ$ resolution over 1981--2019. \textbf{a--c}, Grid-cell Pearson temporal correlation calculated directly from the annual crop-season GPP, ET, and end-of-season yield series, respectively. Rain-fed and irrigated outputs are combined using fixed-2015 crop area. Panel annotations give the median across valid grid cells.}\label{fig_EU_wheat_pretrain}
\end{figure}

We characterize the temporal agreement between the default and pre-trained simulations at 0.5$^\circ$ resolution over 1981--2019. To evaluate the pre-trained initialization, we calculate, for each grid cell, the Pearson correlation directly from the annual series of crop-season GPP, ET, and end-of-season yield. Rain-fed and irrigated outputs are combined using fixed-2015 crop area. Median correlations are 0.82 for GPP, 0.89 for ET, and 0.85 for yield (Fig.~\ref{fig_EU_wheat_pretrain}), showing that the pre-trained simulations provide a process-informed initialization that retains much of the default interannual variability. Pre-training is used only to initialize subsequent fine-tuning and these correlations are not an evaluation of the final model against observations.

\subsection{US maize}

\subsubsection{County-level crop statistics} \label{sec:county-level_data}

The US Corn Belt domain comprises Illinois, Indiana, Iowa, Kansas, Michigan, Minnesota, Missouri, Nebraska, North Dakota, Ohio, South Dakota, and Wisconsin. County production, harvested area, and yield statistics are obtained from USDA-NASS. The 1981--2019 analysis retains 1,047 counties. Reported maize yield is converted to dry matter using Eq.~\ref{eq:standardized_yield} with $h=0.155$.

\subsubsection{Pre-training}

In the US Corn Belt, we pre-train NeuralCrop against outputs generated by its process-based simulations from 2016-1-1 to 2016-12-31 for corn under rain-fed and irrigated management. The training setup is described in Sections~\ref{sec:data} and~\ref{sec:training}.

\begin{figure}[H]
\centering
\includegraphics[width=1.0\textwidth]{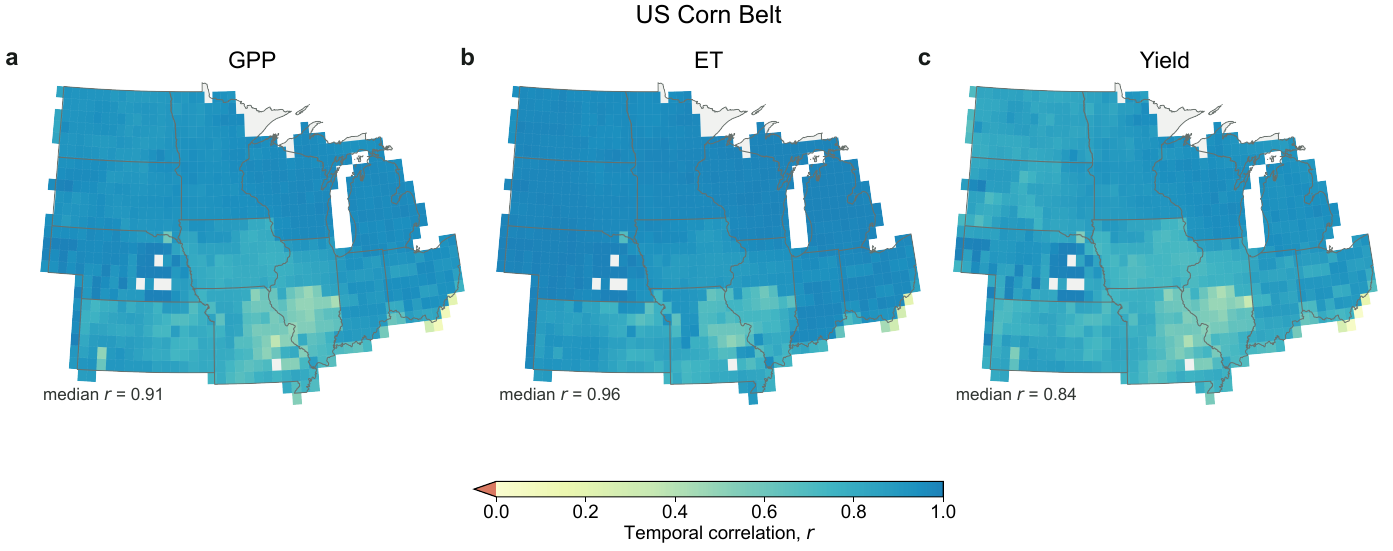}
\caption{Spatial distribution of grid-cell temporal correlations between the default and pre-trained simulations for US Corn Belt at 0.5$^\circ$ resolution over 1981--2019. \textbf{a--c}, Grid-cell Pearson temporal correlation calculated directly from the annual crop-season GPP, ET, and end-of-season yield series, respectively. Rain-fed and irrigated outputs are combined using fixed-2015 crop area. Panel annotations give the median across valid grid cells.}\label{fig_us_corn_pretrain}
\end{figure}

We apply the same 1981--2019 grid-cell comparison to the US Corn Belt. Median correlations are 0.91 for GPP, 0.96 for ET, and 0.84 for yield (Fig.~\ref{fig_us_corn_pretrain}). Thus, the pre-trained maize configuration provides an initialization that retains much of the default interannual variability. Final model performance is evaluated only after fine-tuning.

\subsection{Regional error magnitude}

\begin{figure}[H]
\centering
\includegraphics[width=1\textwidth]{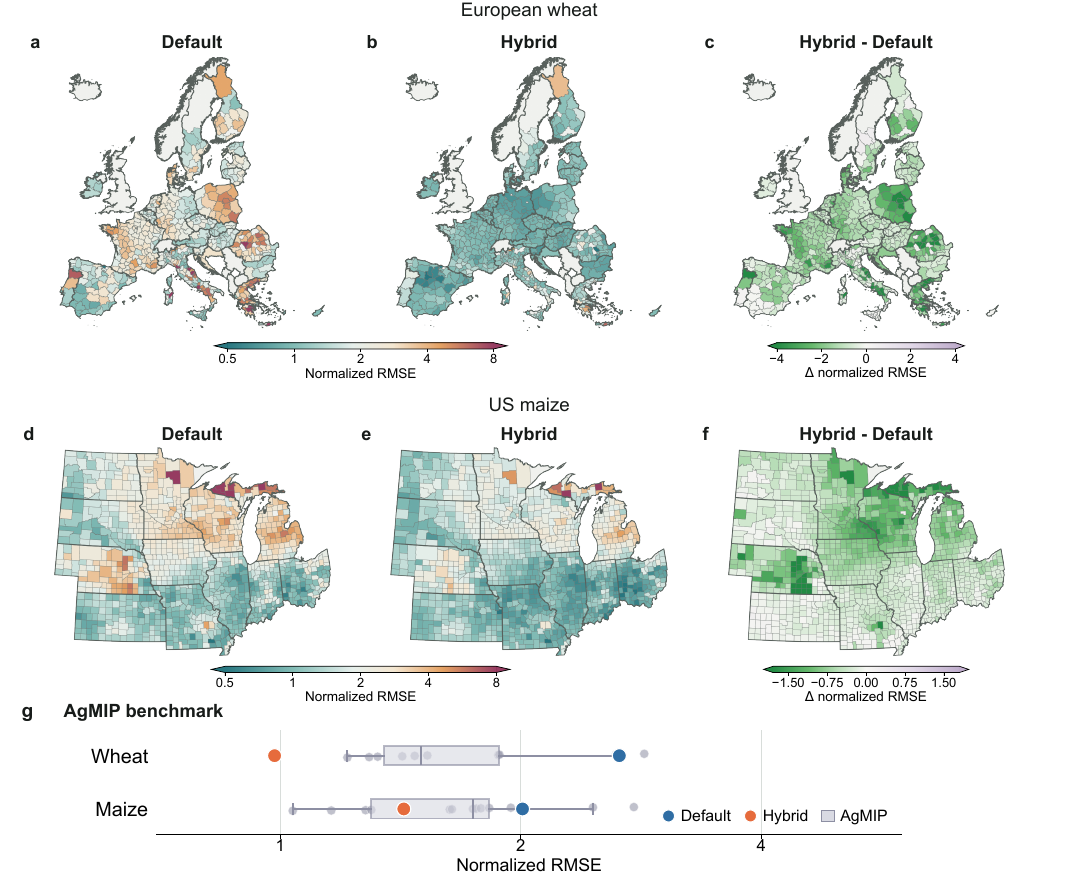}
\caption{Evaluation of yield simulations and comparison with AgMIP models. \textbf{a,b}, Administrative-unit normalized RMSE (nRMSE) between reported and simulated European wheat yields for the default and hybrid simulations over 1981--2019. \textbf{c}, Difference between the nRMSE obtained with the hybrid and fully process-based (default) model versions, respectively ($\Delta nRMSE$). Negative values indicate improvement by the hybrid version and occur across 98.2\% of mapped wheat area. \textbf{d,e}, Corresponding county-level default and hybrid simulations for US Corn Belt maize over 1981--2019. \textbf{f}, Corresponding $\Delta nRMSE$ map. Negative values occur across 98.4\% of mapped maize area. \textbf{g}, Crop area-weighted nRMSE for the default and hybrid simulations and the AgMIP GGCM distributions over the available 1981--2016 period. Points within the AgMIP boxplots denote individual GGCMs.}
\label{fig:regional_yield_nrmse}
\end{figure}

\clearpage
\subsection{Yield skill by distance}

\begin{figure}[H]
\centering
\includegraphics[width=0.96\textwidth]{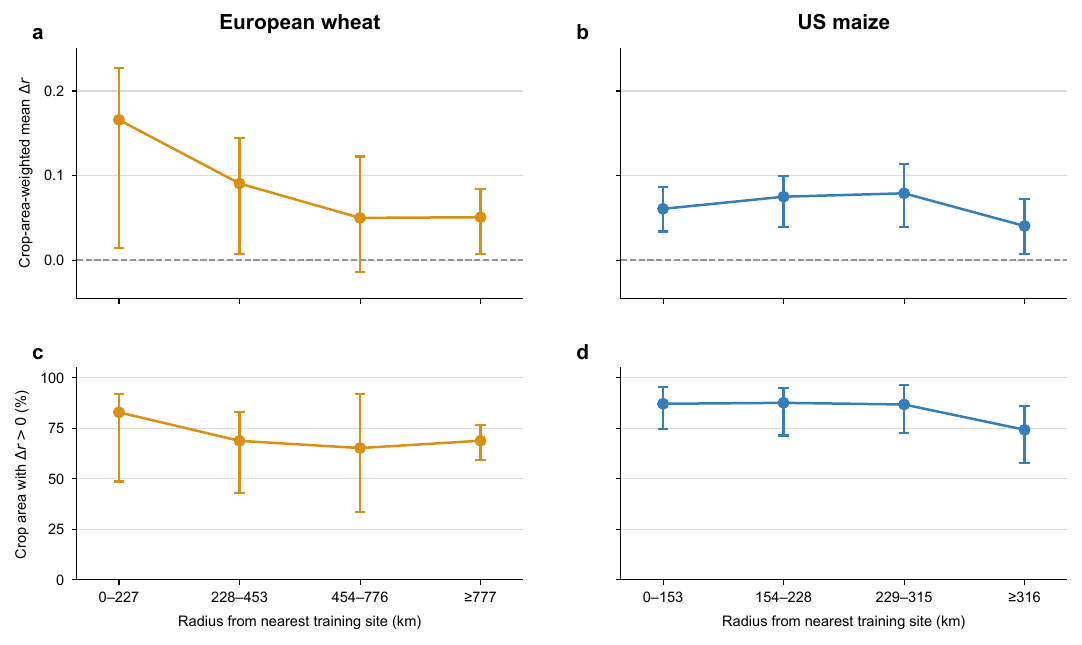}
\caption{Regional yield-correlation gains by radial distance from flux-tower training sites over 1981--2019. Geodesic distances are measured from administrative-unit polygon centroids to the nearest crop-specific training-site model grid-cell center. Each distance class contains approximately one-quarter of the mapped crop area. \textbf{a,b}, Crop area-weighted mean change in administrative-unit temporal correlation ($\Delta r$) for European wheat and US maize. \textbf{c,d}, Mapped crop area with positive $\Delta r$ in the same distance classes. Points show the estimates. Bars show 95\% intervals from 5,000 country-cluster (wheat) or state-cluster (maize) bootstrap resamples. The distance classes differ between crops because their training-site distributions and mapped crop areas differ.}
\label{fig:yield_site_distance}
\end{figure}

We assessed changes in regional yield correlation at different distances from the flux towers used for fine-tuning. For each administrative unit, we calculated the geodesic distance from its polygon centroid to the nearest crop-specific training site, represented by the centre of its $0.5^\circ$ model grid cell. Within each crop, we divided administrative units into four distance classes, each containing approximately one-quarter of the mapped crop area. For each class, we calculated the crop area-weighted mean change in temporal correlation ($\Delta r=r_{\mathrm{hybrid}}-r_{\mathrm{default}}$) and the share of mapped crop area with $\Delta r>0$. We obtained 95\% bootstrap confidence intervals from 5,000 resamples of whole countries for wheat or whole states for maize. Mean $\Delta r$ remains positive in the most distant class for both wheat (+0.051 [0.008, 0.084]) and maize (+0.041 [0.007, 0.072]; Fig.~\ref{fig:yield_site_distance}).

\subsection{Random seed sensitivity}

\begin{figure}[H]
\centering
\includegraphics[width=0.94\textwidth]{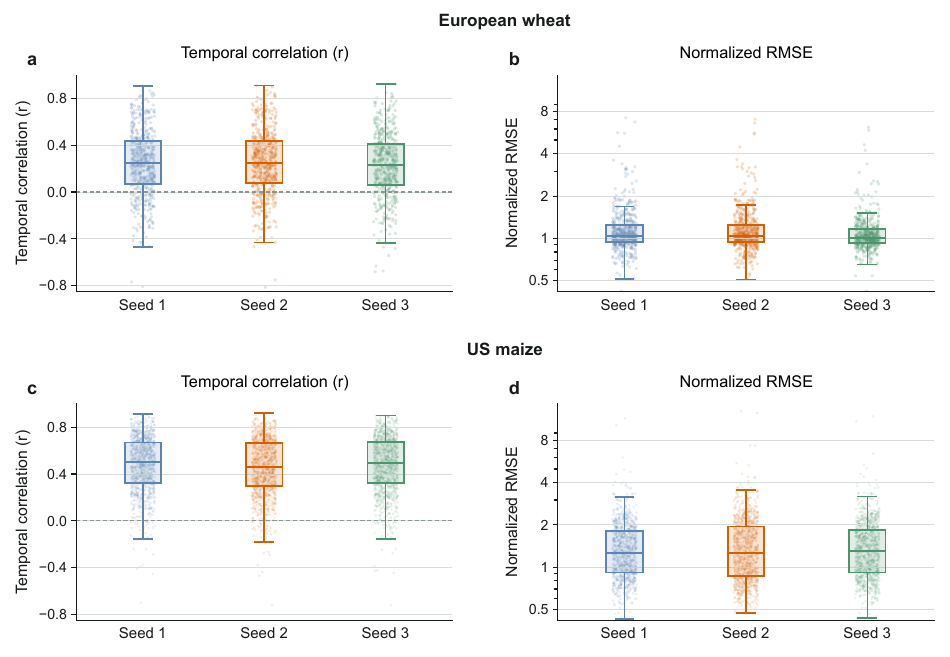}
\caption{Regional yield sensitivity to random seed over 1981--2019. \textbf{a,b}, Distributions of administrative-unit temporal correlation and nRMSE between simulated and reported European wheat yields for three training seeds. \textbf{c,d}, Corresponding distributions for US Corn Belt maize. Points denote administrative units. Boxes show the median and interquartile range and whiskers extend to 1.5 times the interquartile range. nRMSE is displayed on a logarithmic axis.}
\label{fig:yield_seed_sensitivity}
\end{figure}

\subsection{Process transfer diagnostics} \label{sec:process_yield_transfer}

To diagnose transfer from seasonal process signals to yield, we use the selected 1981--2019 regional simulations. For each administrative unit $i$ and process variable $P\in\{\mathrm{GPP},R_{\mathrm{crop}},T_{\mathrm{crop}}\}$, we calculate
\begin{align}
\Delta r_{P,i}
&=r(P^{\mathrm{hybrid}}_i,Y^{\mathrm{obs}}_i)
-r(P^{\mathrm{default}}_i,Y^{\mathrm{obs}}_i),\\
\Delta r_{\mathrm{yield},i}
&=r(Y^{\mathrm{hybrid}}_i,Y^{\mathrm{obs}}_i)
-r(Y^{\mathrm{default}}_i,Y^{\mathrm{obs}}_i),
\end{align}
where $R_{\mathrm{crop}}$ is crop respiration, $T_{\mathrm{crop}}$ is crop transpiration. All annual series are linearly detrended before correlation is calculated. For each crop and process variable, we calculate the crop area-weighted Pearson correlation across administrative units between $\Delta r_{P,i}$ and $\Delta r_{\mathrm{yield},i}$. Confidence intervals are obtained from 5,000 spatial-cluster bootstrap replicates that resample countries for wheat and states for maize.

\begin{figure}[H]
\centering
\includegraphics[width=1.0\textwidth]{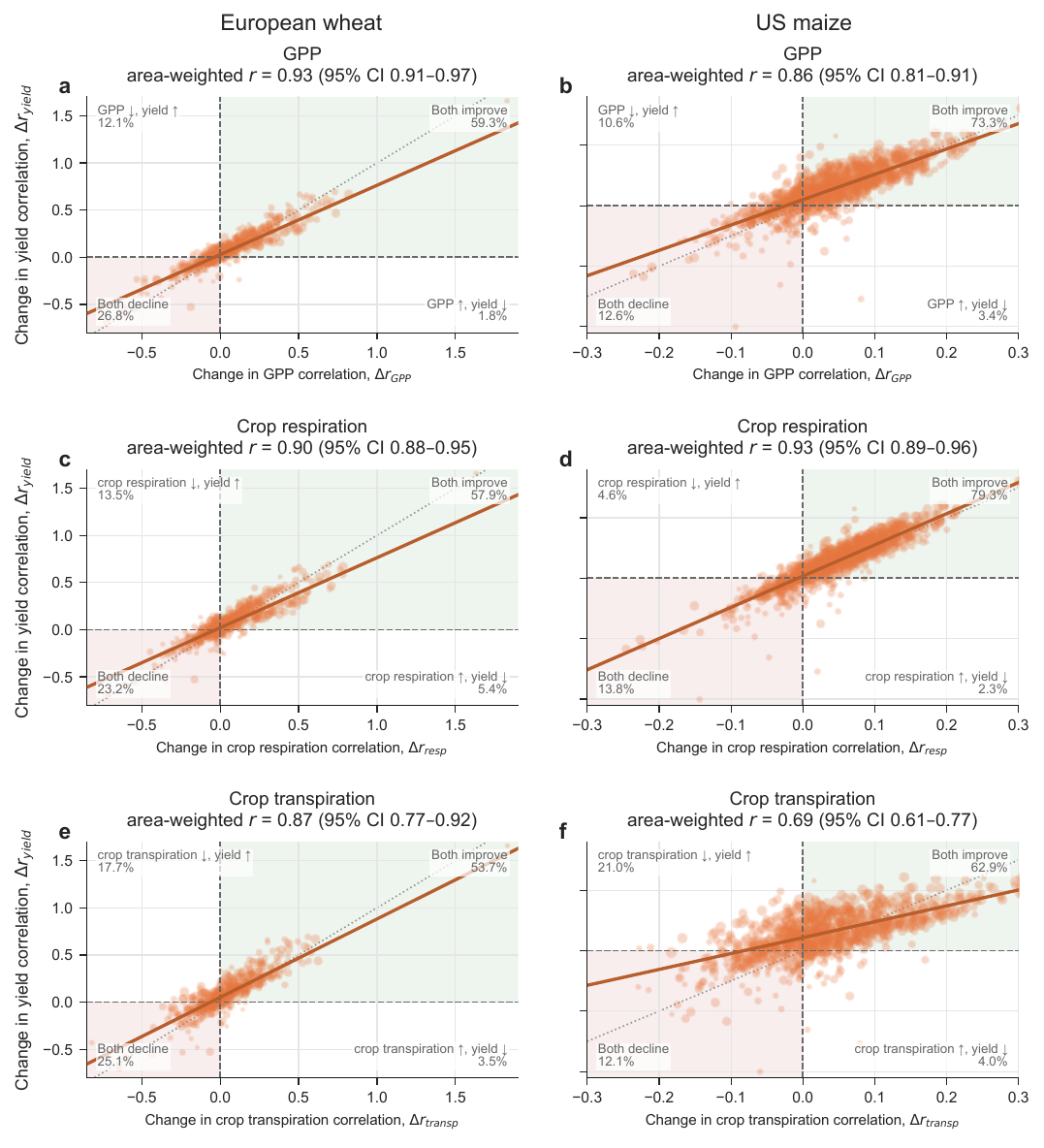}
\caption{Association between changes in seasonal process--yield correlations. \textbf{a,b}, Gross primary productivity (GPP), \textbf{c,d}, crop respiration, and \textbf{e,f}, crop transpiration. Left panels show European wheat and right panels show US Corn Belt maize over 1981--2019. Each point represents an administrative unit. Marker size increases with mapped crop area. Horizontal and vertical axes show changes from the default to the hybrid simulation in temporal correlation with reported yield for the simulated process variable and simulated yield, respectively. Annual process, simulated yield, and reported yield series are linearly detrended separately before correlations are calculated. Crop respiration and transpiration are summed over the active crop season. Solid orange lines show crop area-weighted linear fits. Dotted lines show the 1:1 relationship. Quadrant percentages give mapped crop-area shares. The 95\% confidence intervals for the area-weighted correlations use 5,000 spatial-cluster bootstrap resamples of countries for wheat and states for maize.}
\label{fig:process_yield_transfer}
\end{figure}

\subsection{Soil moisture classification} \label{sec:soil_moisture_classes}

We use monthly ERA5-Land volumetric soil water from the upper three layers (0--7, 7--28, and 28--100 cm) to characterize root-zone soil moisture. The layers are combined using thickness weights of 0.07, 0.21, and 0.72. For each $0.5^\circ$ grid cell and year, we average the monthly values over April--June for European wheat and May--July for US maize, weighting each month by its number of days. Each grid-cell seasonal value is converted to an empirical percentile relative to 1981--2010 values for the same cell. We then average grid-cell percentiles within each administrative unit using crop-area weights, retaining years with at least 90\% crop-area coverage. We interpret this spatial average as a relative soil-moisture index on a 0--100 scale. The five classes are wettest ($\geq80$), wet (60--80), near normal (40--60), dry (20--40), and driest ($<20$). This classification is descriptive and does not by itself identify a causal drought or waterlogging effect. Within each moisture class, we calculate the absolute difference between observed and simulated yield anomalies after standardizing each detrended series by its own interannual standard deviation. We average these errors using mapped crop-area weights. The evaluation covers 1981--2019 for the default and hybrid simulations and the available 1981--2016 Phase 3 period for AgMIP.

Table~\ref{table:moisture_year_support} summarizes the calendar-year support of the five moisture classes and the ten 10-point index bins in main text. A year contributes to a bin when at least one administrative unit falls in it. This does not mean that the entire region experienced that moisture condition. The crop area-weight shares show whether a few years dominate a bin despite a larger number of represented years.

\begin{table}[ht]
\centering
\small
\caption{Calendar-year support and concentration of crop-area weights in the soil-moisture index bins. The count $n$ is the number of distinct calendar years with at least one administrative-unit year in the bin. The leading and top-three year shares are percentages of the total crop-area weight summed over administrative-unit years in that bin. Counts and shares use records over 1981--2019.}
\label{table:moisture_year_support}
\setlength{\tabcolsep}{3pt}
\begin{tabular}{lrrr rrr}
\toprule
& \multicolumn{3}{c}{European wheat} & \multicolumn{3}{c}{US maize}\\
Index interval & $n$ & Leading year (\%) & Top three (\%) & $n$ & Leading year (\%) & Top three (\%)\\
\midrule
\multicolumn{7}{l}{20-point index classes}\\
$<20$ & 35 & 2011 (11.2) & 29.2 & 37 & 1988 (13.7) & 35.7\\
20--40 & 38 & 2015 (8.0) & 23.2 & 39 & 1994 (8.3) & 20.5\\
40--60 & 39 & 2008 (8.0) & 20.8 & 39 & 2002 (6.2) & 16.5\\
60--80 & 38 & 2001 (10.3) & 22.8 & 39 & 1984 (6.5) & 18.5\\
$\geq80$ & 38 & 2013 (18.3) & 43.4 & 38 & 2019 (7.9) & 21.8\\
\midrule
\multicolumn{7}{l}{10-point index bins}\\
0--10 & 30 & 2011 (17.8) & 44.6 & 32 & 1988 (23.7) & 50.8\\
10--20 & 35 & 2012 (9.1) & 26.3 & 37 & 2007 (10.4) & 30.3\\
20--30 & 36 & 2015 (9.8) & 27.1 & 39 & 1994 (9.9) & 22.8\\
30--40 & 37 & 2002 (6.6) & 19.7 & 39 & 2016 (7.4) & 20.2\\
40--50 & 38 & 2008 (10.4) & 24.4 & 39 & 2002 (6.3) & 17.1\\
50--60 & 38 & 2004 (6.6) & 19.0 & 39 & 2002 (6.1) & 17.4\\
60--70 & 38 & 2001 (11.0) & 23.3 & 39 & 2008 (7.0) & 20.2\\
70--80 & 38 & 2001 (9.7) & 24.8 & 39 & 1996 (7.6) & 21.7\\
80--90 & 37 & 2013 (13.0) & 31.0 & 38 & 1995 (7.7) & 21.1\\
90--100 & 31 & 2013 (25.5) & 60.6 & 34 & 1993 (12.8) & 29.0\\
\bottomrule
\end{tabular}
\end{table}

\begin{figure}[H]
\centering
\includegraphics[width=1.0\textwidth]{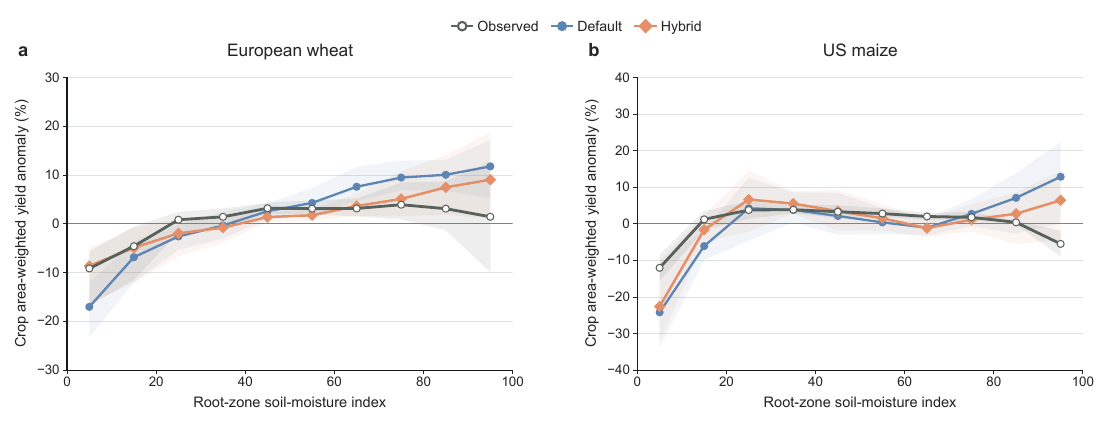}
\caption{Sensitivity of the soil moisture response to anomaly scaling. Yield anomalies are expressed as percentages of the fitted linear trend for each administrative unit. Crop area-weighted means are then calculated in ten 10-point bins of the growing-season root-zone soil-moisture index. Panels \textbf{a} and \textbf{b} show European wheat and US maize, respectively, with curves for observations and the default and hybrid simulations. The shading shows 95\% intervals from 5,000 country- or state-cluster bootstrap resamples.}
\label{fig:moisture_response_percent}
\end{figure}

Figure~\ref{fig:moisture_response_percent} presents moisture-response curves for detrended yield anomalies expressed as percentages of the fitted linear yield trend for each administrative unit. For the response curves, administrative-unit years are assigned to ten 10-point bins of the same soil-moisture index. Within each bin, detrended yield anomalies are averaged using mapped crop area as the weight. We obtain 95\% intervals from 5,000 spatial-cluster bootstrap resamples, sampling countries for European wheat ($n=26$) and states for US maize ($n=12$). This scaling check examines whether the flatter hybrid response can be explained solely by its lower absolute yield level.

\clearpage

\putbib[sn-bibliography-sup]
\end{bibunit}

\end{document}